\documentclass{article}

\PassOptionsToPackage{sort}{natbib}

 \usepackage[preprint]{neurips_2026}

\usepackage[utf8]{inputenc} % allow utf-8 input
\usepackage[T1]{fontenc}    % use 8-bit T1 fonts
\usepackage{hyperref}       % hyperlinks
\usepackage{url}            % simple URL typesetting
\usepackage{booktabs}       % professional-quality tables
\usepackage{amsfonts}       % blackboard math symbols
\usepackage{nicefrac}       % compact symbols for 1/2, etc.
\usepackage{microtype}      % microtypography
\usepackage{xcolor}         % colors
\usepackage{graphicx}
\usepackage{subcaption}
\usepackage{tabularx}
\usepackage{array}
\usepackage{ragged2e}
\usepackage{siunitx}
\usepackage[table]{xcolor}
\usepackage[most]{tcolorbox}
\tcbuselibrary{breakable,skins,listings}
\usepackage{float}
\usepackage{enumitem}
\setlist[itemize]{leftmargin=2em}
\setlist[enumerate]{leftmargin=2em}
\usepackage{soul}
\usepackage{multirow}

\definecolor{grayblue}{HTML}{6E8898}
\hypersetup{
    colorlinks=true,      
    linkcolor=grayblue,       
    citecolor=grayblue,       
    urlcolor=grayblue         
}

\bibpunct{(}{)}{;}{a}{,}{,}

\newcolumntype{Y}{>{\RaggedRight\arraybackslash}X}
\usepackage{listings}
\usepackage{inconsolata}
\usepackage{amsmath}
\usepackage{amssymb}
\usepackage{mathtools}
\usepackage{amsthm}
\usepackage{bbm}

\usepackage[capitalise,noabbrev]{cleveref}
\newcommand{\appref}[1]{\hyperref[#1]{Appendix~\ref*{#1}}}
\newcommand{\secref}[1]{\hyperref[#1]{$\S$\ref*{#1}}}

\theoremstyle{plain}

\usepackage{colortbl}

\theoremstyle{definition}

\theoremstyle{remark}

\title{AI Assistants Overassist}

\author{%
  Verona Teo$^{1*}$\footnotetext[1]{Equal contribution} \qquad
  Raghav Jain$^{2*}$ \qquad
  Tobias Gerstenberg$^{1}$ \qquad
  Max Kleiman-Weiner$^{3}$ \\[1.5ex]
  \normalfont
  $^1$Stanford University \\
  $^2$University of California, San Diego \\
  $^3$University of Washington \\
}

\begin{document}

\addtocontents{toc}{\protect\setcounter{tocdepth}{-1}}

\maketitle
{\renewcommand{\thefootnote}{\fnsymbol{footnote}}\footnotetext[1]{Equal contribution}}

\begin{abstract}

Large language models (LLMs) are increasingly used as tutors and thought partners, helping users reason through problems.
While guidance from AI assistants can scaffold thinking and foster learning, such benefits depend on \emph{how} they help---for instance, intervening too early or too frequently may hinder true learning and cognitive engagement.
Yet how AI systems navigate intervention decisions during problem-solving remains poorly understood.
Here, we introduce \textsc{Int-Bench}, a simulation-based benchmark for evaluating LLM interventions during learning.
\textsc{Int-Bench} simulates a ``student'' solving a problem while a ``teacher'' monitors the student's reasoning and decides whether, when, and how to intervene.
Across three domains---code debugging, mathematics, and brain teasers---we evaluate LLM teachers on the frequency and timing of interventions, as well as their impact on both immediate task success and generalization to new problems. 
We also compare LLMs to humans, finding that LLMs intervene more frequently and earlier than humans. 
Moreover, in contrast to humans, they tend to provide complete solutions rather than targeted hints.
These findings suggest that current LLM assistants often optimize for short-term success rather than supporting the reasoning processes needed for deeper learning and long-term success.

\end{abstract}

\section{Introduction}

Large language models (LLMs) are being adopted in a wide range of educational and professional workflows \citep{chatterji2025people, peng2023impact}. 
In these settings, the value of such assistance depends not only on whether the model can solve the task, but also on \textit{how} it supports the user's reasoning along the way. 
Effective help requires deciding both when to intervene and how much information to provide, as well as when to stay silent and let a learner reason on their own.
Intervening too early or too directly can take over the reasoning process, while waiting too long may leave users stuck \citep{soderstrom2015learning, mclaren2014exploring}. 
This reflects a trade-off in learning environments, where teachers and parents balance the benefits of intervention and efficient task completion against the longer-term value of productive struggle and perseverance \citep{campbell2025experimental, shachnai2025bayesians}. 

Recent work suggests that while AI assistance can improve immediate task performance, it can negatively affect cognitive engagement, motivation, and learning \citep{shaw2026thinking, faas2024give, he2025plan}. 
% Across different tasks, prior work shows that AI assistance can reduce cognitive load and improve immediate performance, but assistance may also reduce autonomy and cognitive engagement \citep{kosmyna2025brainchatgptaccumulationcognitive, stadler2024cognitive, chen2025more, kumar2025human, liu2026ai}. 
However, these studies primarily measure downstream outcomes of assistance.
Less is known about the assistance behavior that may give rise to these detrimental effects: when LLMs choose to intervene, how early they step in, how much information they provide, and whether their feedback supports reasoning or simply moves the user toward the answer. 

Motivated by this gap, we study LLM assistance in a controlled, simulated student-teacher setting, allowing us to characterize LLM intervention behavior and compare it with human assistance strategies. 
We address the following research questions:

\begin{enumerate}[label=\textbf{RQ\arabic*:}, leftmargin=*, labelindent=1em]
    \item How frequently do LLMs intervene, and at what point in the problem-solving process do these interventions occur?
    \item To what extent do LLM interventions improve (or hurt) immediate task success?
    \item What is the effect of LLM interventions on the student's ability to generalize to related, unseen problems?
    \item How does the intervention behavior of LLMs differ from that of humans in similar settings? 
\end{enumerate}

\begin{figure*}[t]
  \centering
  \includegraphics[width=\linewidth]{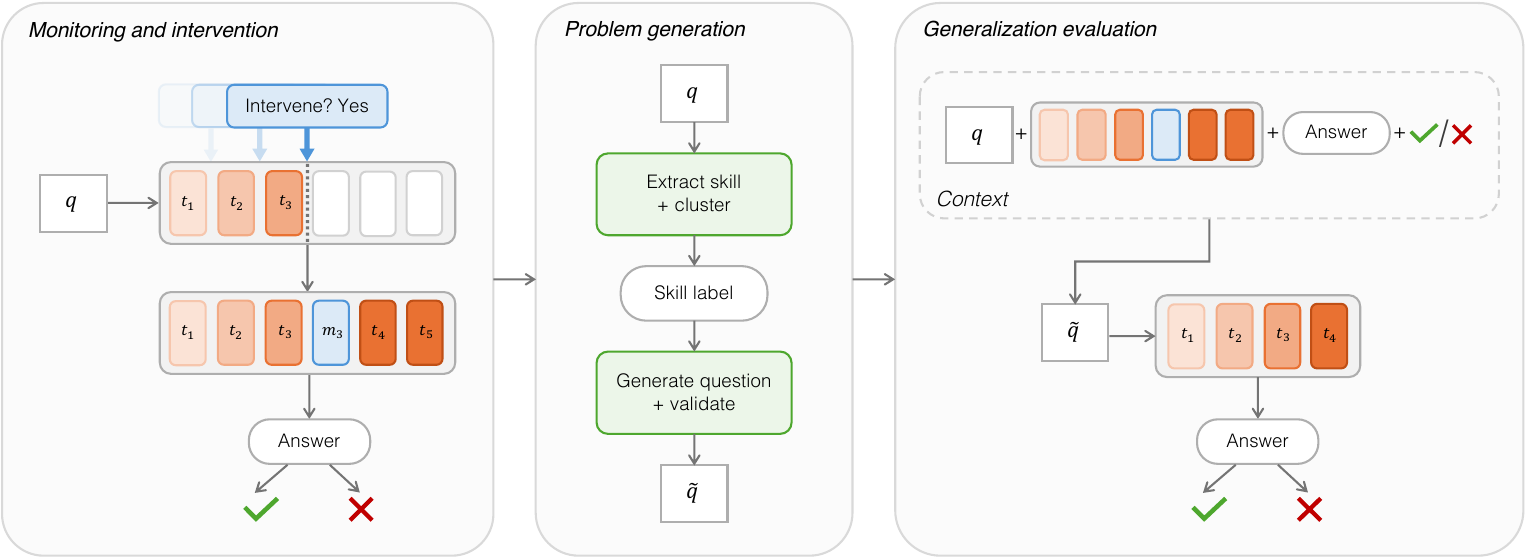}
  \caption{\textbf{Overview of \textsc{Int-Bench}.} For a question $q$, the student produces a baseline reasoning trace $t = (t_1, t_2, \dots, t_T)$. 
   If the teacher decides to intervene at step $k$, an intervention message $m_{k}$ is injected into the reasoning trace.
  The student then updates its reasoning and provides an answer, which a judge evaluates for correctness. 
  A structured generation module produces a related new problem $\tilde{q}$ based on the original question $q$'s skills. 
  For this generalization task, the student receives the original problem, reasoning trace, message, answer, and correctness verdict, which serves as the context when attempting the new question $\tilde{q}$.}

  \label{fig:intro}
\end{figure*}

Our work makes three main contributions. 
First, we formalize LLM assistance as a sequential intervention game and introduce \textsc{Int-Bench} (\autoref{fig:intro}), a simulation-based benchmark where a teacher LLM monitors a student's reasoning trace and decides whether, when, and how to intervene. 
Second, we develop metrics for characterizing assistance behavior, including intervention frequency, timing, immediate helpfulness, and generalization to new, related problems. 
We use these metrics in a large-scale empirical analysis across multiple models and domains. 
Third, we conduct a human study where participants act as teachers in similar settings, enabling direct comparisons between LLMs and humans.
Overall, we find that LLMs intervene more frequently and earlier than humans, often providing overly informative feedback that mostly gives the solution away. 
Furthermore, these interventions tend to be highly problem-specific, limiting the student's ability to generalize and apply what was learned to new problems.

\section{Related Work}

\paragraph{Impacts of AI on learning}

AI assistance can improve immediate performance while reducing cognitive engagement and autonomy \citep{stadler2024cognitive, chen2025more, faas2024give, kosmyna2025brainchatgptaccumulationcognitive}.
More broadly, delegating mental work to external tools can erode independent analytical capacity over time \citep{risko2016cognitive, liu2026ai}.
Making learning effortful, even at the cost of initial performance, can improve long-term retention and transfer \citep{bjork1994memory, kapur2014productive, gajos2022dopeopleengage}.
These studies primarily focus on the downstream \emph{outcomes} of assistance;
our work additionally characterizes the assistance behavior that produces them, including when models intervene, how much they reveal, and whether their feedback supports reasoning or substitutes for it.

\paragraph{Timing and boundaries of AI assistance}

The tension between providing support and withholding assistance to promote learning---often known as the \emph{assistance dilemma}---is a core challenge in educational science \citep{koedinger2007exploring, mclaren2008whenandhow, mclaren2014exploring, maniktala2020extendingthehint}.
Recent work has begun operationalizing similar trade-offs in human-AI collaboration. 
In particular, several systems have focused on learning \emph{when} an agent should speak or stay silent \citep{steyvers2025helpplanninglastinghumanai, patel2026discussllm, manzoor2025learning}.
However, these approaches typically operate at the level of discrete decision instances or dialogue turns.
We instead formalize assistance as sequential monitoring of a reasoning trace, allowing for finer-grained control over intervention timing within a single problem, and evaluate whether interventions promote learning transfer.

\paragraph{Simulated students and teachers}
% LLMs are increasingly deployed as tutors and pedagogical feedback systems,
Prior work has explored LLMs' abilities to provide feedback and produce appropriate teacher responses \citep{tack2022ai, macina2023mathdial, matelsky2023largelanguagemodelassistededucation}.
LLM-simulated students have been used to study learning processes \citep{ross2025modelingstudentlearning38, ross2025learningmakemistakesmodeling} and to support teacher training \citep{markel2023gpteach, abbasiantaeb2023letllmstalksimulating, lu2024generativestudents, liu2024socraticlm, jin2025teachtunereviewingpedagogicalagents, daheim-etal-2024-stepwise, hu2025exploringthepotential}.
We build on this line of work by simulating both students and teachers using LLMs in an intervention game.

\section{The \textsc{Int-Bench} Framework}
\label{sec:framework}

\subsection{Problem Setup}
\label{game}

We formalize the interaction between a student and a teacher as a sequential intervention game (\autoref{fig:intro}).
Let $\mathcal{Q} = \{q_1, \ldots, q_N\}$ be a set of $N$ question instances. 
Each episode of the game involves four components: a problem $q_i \in \mathcal{Q}$, a student model, a teacher model, and a judge. 
An episode proceeds in two phases. 

In the \textit{baseline} phase, the student first solves $q$ without assistance, generating a baseline reasoning trajectory $t = (t_1, t_2, \dots, t_T)$ and a final answer $\hat{y}$, which is evaluated by the judge against a reference solution $y^*$. 
In the \emph{monitoring} phase, the baseline reasoning trace is revealed to the teacher, who decides whether to intervene.
At each step $k < T$, the teacher observes the reasoning prefix $t_{1:k}=(t_1, \dots, t_k)$ and selects an action $a_k \in \{\text{wait}, \text{intervene}\}$.
A teacher policy $\pi$ maps the observed prefix to an action and, if intervening, a message: $\pi(t_{1:k}) = (a_k, m_k)$.
If $a_k = \text{intervene}$, the message $m_k$ is immediately injected into the trace after $t_{1:k}$, the monitoring phase ends, and the student updates its reasoning given the intervention.

\subsection{Monitoring}
\label{sec:monitoring}
% During the monitored rollout, the teacher observes the student's baseline reasoning trace.
We compare teacher behavior during the monitoring phase under two conditions.
In the \emph{Standard} condition, the baseline trajectory is revealed to the teacher in cumulative increments of a fixed size $s$ (e.g., characters).
At each step $k$, the teacher observes the reasoning prefix $t_{1:k}$ and chooses between waiting (reveals next increment) or intervening (generates an intervention message $m_k$ and monitoring terminates). 
The teacher may intervene at most once.

We also evaluate an \emph{Oracle} condition, in which the teacher receives the full baseline trace $t$, the student's final answer $\hat{y}$, and the  correctness verdict $\mathbbm{1}[\hat{y} = y^*]$ simultaneously, prior to making any decision.
The teacher decides whether to intervene, and if so, selects the optimal point $k$ at which to intervene post hoc.

We define two behavioral metrics:

\paragraph{Intervention Frequency}
The proportion of episodes where the teacher assists, $\phi = \frac{1}{N}\sum_{i=1}^N I_i$, where $I_i \in \{0,1\}$ indicates an intervention on problem $q_i$. 
We additionally compute conditional frequencies 
% $\phi_{\text{correct}}$ and $\phi_{\text{incorrect}}$ based on the student's unassisted baseline answer $\hat{y}_i$, where
$\phi_{\text{correct}} = \mathbbm{P}(I_i=1 \mid \mathbbm{1}[\hat{y}_i = y^*_i]=1)$ and $\phi_{\text{incorrect}} = \mathbbm{P}(I_i=1 \mid \mathbbm{1}[\hat{y}_i = y^*_i]=0)$, which capture how often teachers intervene when the student would have succeeded or failed without assistance.

\paragraph{Intervention Timing}
We define the \emph{absolute timing} $\tau_{\text{abs}}$ as the length of the reasoning trace (e.g., number of characters) revealed before the intervention. 
The \emph{relative timing} $\tau_{\text{rel}}$ is the normalized point of intervention, defined as $\tau_{\text{rel}}= \tau_{\text{abs}}/L \in [0,1]$, where $L$ is the total length of the baseline trajectory. 
Values near $0$ indicate early intervention.

\begin{figure}
    \centering
    \includegraphics[width=\linewidth]{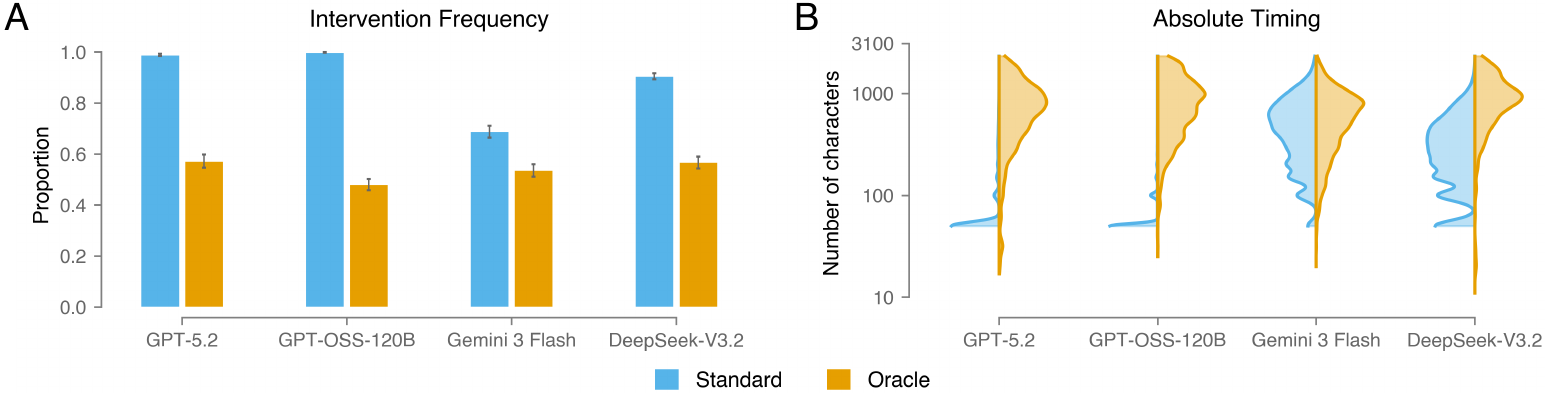}
    \caption{\textbf{Intervention frequency and absolute timing for each model across all three domains.} 
    \textbf{(A)} Across all models, teachers in the \textit{Standard} condition intervened more often than \textit{Oracle} teachers. Error bars are 95\% bootstrapped CIs.
    \textbf{(B)} Number of characters that teachers chose to reveal from the reasoning trace prior to intervening. 
    \textit{Standard} teachers received 50-character increments at a time.}
    \label{fig:frequency-timing}
\end{figure}

\subsection{Post-Intervention Reasoning}
\label{sec:post-intervention}
Following an intervention at step $k$, we construct an updated context by truncating the baseline
reasoning trace to $t_{1:k}$ and appending the teacher's intervention message $m_k$. 
The student then attempts to solve the problem using one of three update strategies:

\begin{enumerate}
    \item{\emph{Standard-Continue}:} The student receives feedback from the \textit{Standard} teacher and generates a revised reasoning trajectory  $t'$ and a final answer $\hat{y}'$ conditioned on the updated context.
    \item{\emph{Oracle-Continue}:} The student receives feedback from the \textit{Oracle} teacher and similarly generates a revised trajectory $t'$ and a final answer $\hat{y}'$.
    \item{\emph{Stop-and-Answer}:} The student receives feedback from the \textit{Standard} teacher, but is prevented from generating any further reasoning steps, and must output a final answer $\hat{y}'$ immediately.
\end{enumerate}

To quantify the impact of these strategies on task performance, we measure the average signed change in answer correctness following an intervention across all intervened episodes.

\paragraph{Immediate Helpfulness}
Let $\mathcal{I}$ denote the set of episodes where the teacher intervenes.
For each $i \in \mathcal{I}$, we define $H_i = \mathbbm{1}[\hat{y}'_i = y^*_i] - \mathbbm{1}[\hat{y}_i = y^*_i] \in \{-1, 0, 1\}$, where $\hat{y}_i$ and $\hat{y}'_i$ are the baseline and post-intervention answers, respectively. 
The overall immediate helpfulness score $H = \frac{1}{|\mathcal{I}|}\sum_{i \in \mathcal{I}} H_i$ reflects the net accuracy gain across intervened episodes.

\subsection{Problem Generation}
\label{sec:generation}
To understand whether interventions promote transfer to new problems, we evaluate students on related but distinct problems that require the same underlying skill as the original questions. 
We use a structured four-step pipeline, adapted from \citet{didolkar2024metacognitive}, to generate variants $\tilde{q}$ for each reference problem $q$.
First, an LLM extracts a fine-grained skill label from $q$ (e.g., ``modular arithmetic'').
Second, these labels are clustered into broader skill categories.
Third, a generator LLM produces candidate variant problems conditioned on $q$ and its skill category, ensuring they require the same underlying skill to solve, but differ in surface form, context, and parameters.
Finally, a validation step filters out candidates that are inconsistent with the target skill category or contain incorrect reference solutions.
A single valid variant $\tilde{q}$ is sampled for the generalization evaluation.

\subsection{Generalization Evaluation}
\label{sec:generalization}
We evaluate the student's performance on the variant problem $\tilde{q}$ under three context conditions to isolate what aspects of prior experience drive generalization ability:
\begin{enumerate}
    \item{\emph{No-Context}:} The student attempts to solve $\tilde{q}$ from scratch without any information about the original problem $q$. 
    This serves as a generalization baseline.
    \item{\emph{Problem-Context}:} The student receives the original problem $q$, its unassisted baseline reasoning trace $t$, the final answer $\hat{y}$, and the judge's correctness verdict. 
     This measures whether exposure to a related question and its solution alone helps the student solve the new problem.
    \item{\emph{Intervention-Context}:} The student receives the full assisted episode for $q$ (from the \textit{Standard} monitoring condition), including the reasoning prefix $t_{1:k}$, the teacher's intervention message $m_k$, the post-intervention trace $t'$, the updated answer $\hat{y}'$, and the judge's correctness verdict. 
    This measures the value added by the intervention over simple problem exposure.
\end{enumerate}

In all context conditions, the judge evaluates the student's answer to $\tilde{q}$ against its reference solution.
To isolate the transfer value of these contexts, we define a generalization metric.

\paragraph{Generalization Helpfulness}
The change in correctness on the variant $\tilde{q}_i$ relative to the \textit{No-Context} baseline. 
For each reference-variant pair, $G_i = \mathbbm{1}[\hat{\tilde{y}}^{\text{C}}_i = \tilde{y}^*_i] - \mathbbm{1}[\hat{\tilde{y}}^{\text{NC}}_i = \tilde{y}^*_i]$, where $\hat{\tilde{y}}^{\text{C}}_i$ and  $\hat{\tilde{y}}^{\text{NC}}_i$ are the student's answers with (\textit{Problem-Context} or \textit{Intervention-Context}) and without context, respectively.
The overall generalization helpfulness score $G = \frac{1}{M}\sum_{i=1}^M G_i$ is the average over all $M$ reference-variant pairs.

\begin{figure*}[tb]
    \centering
    \includegraphics[width=\linewidth]{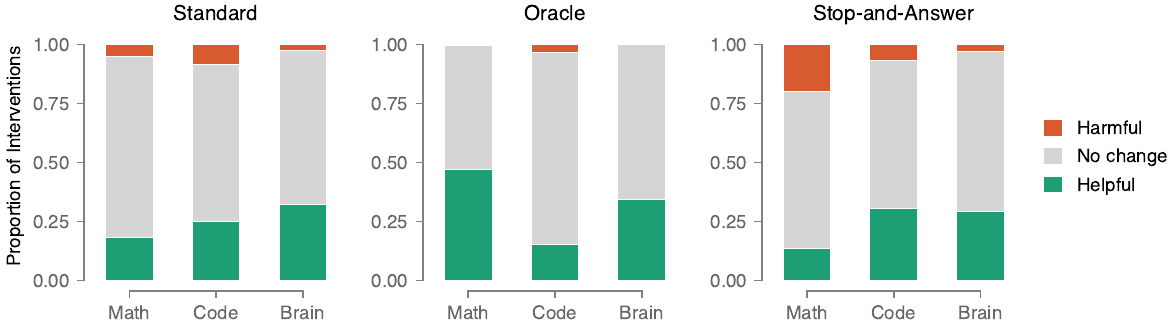}
    \caption{\textbf{Proportion of interventions that were helpful, harmful, or made no difference to the correctness of answers}. Bars are averaged over all teacher models. Helpful interventions are those in which a student's originally incorrect answer became correct after intervention, while harmful interventions are those in which an initially correct answer became incorrect.}
    \label{fig:helpfulness}
\end{figure*}

\section{Experiments}

\subsection{Simulation Setup}

\paragraph{Datasets}

We evaluate our framework on 1,500 problems across three domains:

\begin{itemize}
    \item \textbf{Code Debugging}:
    We sampled 500 code debugging problems from the DebugEval dataset \citep{yang-etal-2025-coast}, which includes buggy code snippets and corresponding solutions.
    
    \item \textbf{Mathematics}:
    We used the MATH-500 dataset \citep{hendrycks2021measuringmathematicalproblemsolving}, a subset of the MATH benchmark designed to evaluate mathematical reasoning. 
    
    \item \textbf{Brain Teasers}:
    We scraped problems from the Braingle website,\footnote{\url{https://www.braingle.com}} an online platform of lateral thinking puzzles.
    We included problems from four different categories (``Riddle,'' ``Language,'' ``Rebus,'' and ``Group'') and sampled the 500 most popular and easiest problems (\appref{braingle-data}).
\end{itemize}

For the generalization evaluation, we first sampled 150 problems per domain to generate variant questions via the pipeline described in \secref{sec:generation}.
We then subsampled 100 successfully validated reference-variant pairs per domain (300 total).

\paragraph{Models}
We used Qwen2.5-7B-Instruct as the student model, which was prompted to show its reasoning step-by-step. 
We selected this model because it exhibits intermediate baseline performance on each of our selected domains, providing sufficient headroom to observe learning from teacher interventions. 
For teacher models, we evaluated two closed-source models (GPT-5.2, Gemini 3 Flash) and two open-source models (GPT-OSS-120B, DeepSeek-V3.2).
We ran each model three times per question.
We used GPT-5.2 as the judge model for evaluating student answers.
All student and teacher models were run with temperature 0.7, and the judge model was run with temperature 0.

For each experiment and the results that follow, we set a fixed increment size of $s=50$ characters (i.e., teachers in the \textit{Standard} monitoring condition received 50-character increments).
\footnote{We also ran experiments with different increment sizes (e.g., fixed 300-character increments, increments at the sentence level), student models (e.g., Qwen3-32B, LLaMA-3.1-8B), and prompt variations (e.g., ``only intervene if truly necessary''). 
We found that these variations resulted in qualitatively similar results; full details are provided in \autoref{sec:additional-experiments}.}

\subsection{Human Studies}
\label{sec:human_exps}
To compare LLM-based interventions with human behavior, we conducted a study with two conditions---\textit{Standard} and \textit{Oracle}---where human participants acted as teachers monitoring simulated student reasoning.
We manually selected 30 problems from the brain teaser dataset and used the same reasoning traces generated by our student model, allowing us to make direct comparisons between human and LLM teachers.

We chose brain teasers because they require minimal domain expertise. 
Unlike math and code debugging, in which teachers need specialized conceptual or technical knowledge to evaluate student reasoning, brain teasers primarily rely on domain-general skills, such as ordinary language understanding and cognitive reflection.
This ensures that variation in participant behavior reflects differences in intervention strategy rather than domain expertise.

We recruited 25 participants for each condition via Prolific, for a total of $N=50$ participants.
Each participant completed 6 trials (episodes), which were randomly ordered.
In the \textit{Standard} condition, participants could reveal the reasoning trace in 50-character increments and could intervene at any point by providing a written message. 
If they reached the end of the trace without intervening, they could either provide feedback or proceed to the next trial.
In the \textit{Oracle} condition, participants were asked whether and when they would intervene only once they had read the entire reasoning trace.

\begin{figure*}[tb]
    \centering
    \includegraphics[width=0.68\linewidth]{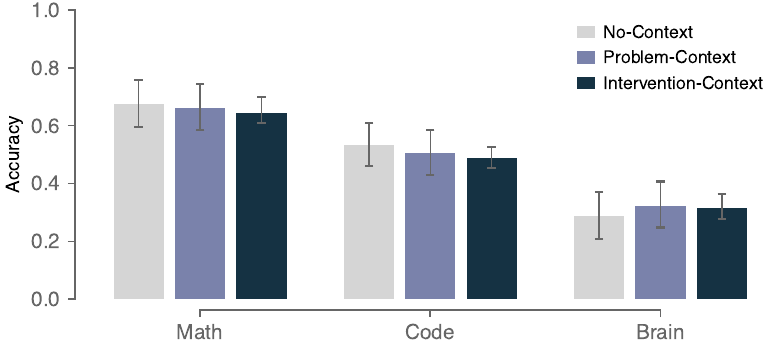}
    \caption{\textbf{Proportion of problems answered correctly on the variant questions.} Bars show the student accuracy under the \textit{No-Context}, \textit{Problem-Context}, and \textit{Intervention-Context} conditions across the math, code debugging, and brain teaser domains. \textit{Intervention-Context} is averaged over the four teacher models. Error bars represent 95\% bootstrapped CIs.}
    \label{fig:generalization}
\end{figure*}

\section{Results}

We organize our results around our four research questions. 
We find that LLMs intervene frequently and early (\secref{sec:rq1}), interventions are moderately helpful for immediate task success (\secref{sec:rq2}), interventions do not reliably improve generalization (\secref{sec:rq3}), and LLM intervention behavior differs both quantitatively and qualitatively from that of humans (\secref{sec:rq4}).

\paragraph{Student baseline accuracy}
Across all three domains, the student achieved a baseline accuracy of 43\% prior to interventions: 70.4\% on math, 45.2\% on code debugging, and 14.4\% on brain teasers.

\subsection{How often and when do LLM teachers intervene? (RQ1)}
\label{sec:rq1}

\paragraph{LLMs intervene frequently and early, even when the student would have gotten the correct answer.}

\autoref{fig:frequency-timing} shows the overall intervention frequency and absolute timing for each model and domain.
We found that in the \textit{Standard} monitoring condition, most models intervened very frequently and early in the reasoning process ($\phi = 0.90$, $\tau_{\text{rel}} = 0.18$).
GPT-5.2, GPT-OSS-120B, and DeepSeek-V3.2 intervened in over 90\% of trials across the three domains, with GPT-5.2 and GPT-OSS-120B teachers intervening even before the second increment (100 characters) had been shown. 
Furthermore, with the exception of Gemini 3 Flash, \textit{Standard} teachers often intervened even when the student would have gotten the answer correct.
For instance, GPT-5.2 and GPT-OSS-120B intervened on questions initially correctly solved 98\% and 100\% of the time, respectively.

\paragraph{Teachers intervene less often when they have full information.}
In the \textit{Oracle} condition, models intervened significantly less frequently compared to \textit{Standard} teachers (\textit{Standard}: $\phi=0.90$, 95\% confidence interval (CI) $[0.89, 0.90]$; \textit{Oracle}: $\phi=0.54$, 95\% CI $[0.52, 0.56]$; $p < .001$). 
They also intervened significantly later in the reasoning trace (\textit{Standard}: $\tau_{\text{rel}} = 0.18$, 95\% CI $[0.17, 0.18]$; \textit{Oracle}: $\tau_{\text{rel}} = 0.56$, 95\% CI $[0.55, 0.57]$; $p < .001$).
Here, the models almost never intervened when the student would have arrived at the correct answer.
In cases where the student would have succeeded on their own, models intervened less than 3\% of the time, indicating that student correctness is a primary factor for their intervention decisions.
% This suggests a relatively ``shallow'' heuristic where the LLM's decision to intervene is primarily driven by the student's final answer correctness.
Compared to other models, Gemini 3 Flash had the smallest difference in the number of characters shown prior to intervening between the \textit{Standard} and \textit{Oracle} conditions.
% These results suggest that models can recognize when (not) to intervene, and in particular, choose not to intervene when the student is able to solve the problem without help.

\subsection{How helpful are the interventions? (RQ2)}
\label{sec:rq2}

\begin{figure}
    \centering
    \includegraphics[width=\linewidth]{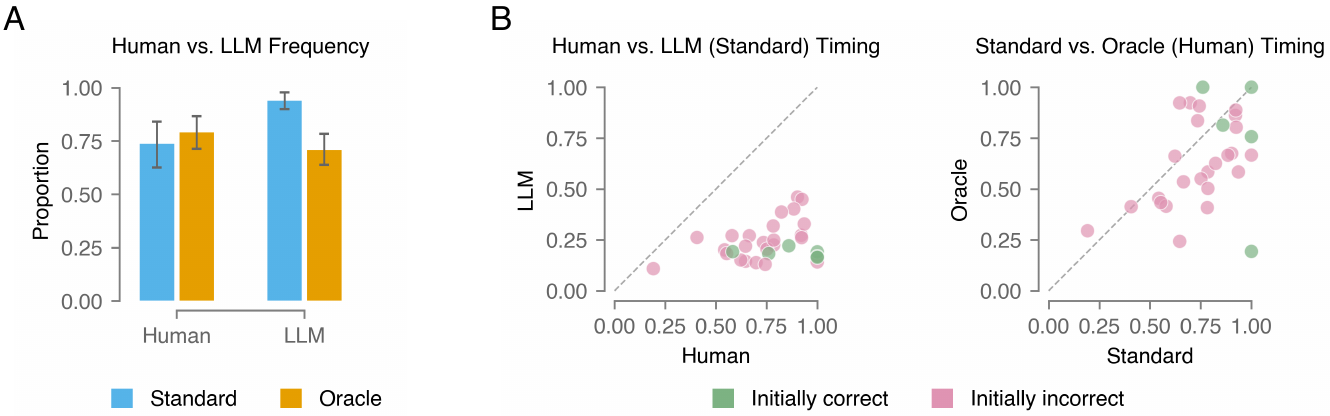}
    \caption{\textbf{Human vs. LLM intervention behavior on brain teasers.}
    \textbf{(A)} Intervention frequency for human and LLM teachers under the \textit{Standard} and \textit{Oracle} conditions. Error bars are bootstrapped 95\% CIs.
    \textbf{(B)} Per-question relative timing for humans vs. LLMs in the \textit{Standard} condition (left) and human relative timing under the \textit{Standard} vs. \textit{Oracle} conditions (right). 
    Humans intervene later than LLMs, including for questions the student would have gotten right without intervention (green dots).
    }
    \label{fig:human-llm}
\end{figure}

\paragraph{Oracle interventions are most helpful (and least harmful).}
Across the domains and models, \textit{Standard} teacher interventions were moderately helpful, resulting in a net accuracy gain of $H=0.20$, with 25.5\% initially incorrect answers becoming correct and 5.4\% becoming incorrect post-intervention (\autoref{fig:helpfulness}).
The interventions from the \textit{Oracle} condition were the most helpful overall ($H=0.30$), particularly in math, and also almost never hurt student performance---the rate at which they caused a correct student to become incorrect was only 1.2\% across all domains. 
So, when models have access to the full reasoning and correctness verdict, they can intervene more effectively.

\paragraph{Interventions are overly informative.}
To understand the extent to which interventions improved student reasoning, we compared against a \textit{Stop-and-Answer} ablated baseline.
In the math domain, we observed that forcing the student to answer without further reasoning after the intervention reduced immediate helpfulness overall compared to the \textit{Standard-Continue} and \textit{Oracle-Continue} conditions.
This suggests that, at least for math, the value of the intervention lies partly in guiding subsequent reasoning; simply stopping the student is not as effective as allowing them to incorporate the feedback.

In contrast, for code debugging problems, the \textit{Stop-and-Answer} condition actually \emph{improved} immediate helpfulness over the \textit{Standard-Continue} condition ($H=0.24$ vs.\ $0.17$), and for brain teasers, the two conditions resulted in similar performance ($H=0.27$ vs.\ $0.30$). 
One possible explanation is that teachers tend to provide long, detailed interventions that often either directly reveal the solution or make the solution easy to infer.
Such interventions mean that students do not need to continue reasoning before arriving at the answer---they can simply read and extract the solution.
We explore this hypothesis further in \secref{sec:rq4}.

\begin{table*}[t]
\centering
\caption{\textbf{Distribution of intervention categories on the 30 brain teaser questions.} Values are the percentage of interventions assigned to each category (each intervention has a single category label) for humans and LLMs under both monitoring conditions.
Subtable (a) characterizes the functional role of the intervention, and (b) measures the extent to which the intervention leaks the solution. Column values sum to 100\% within each subtable.}
\label{tab:intervention-categories}

\setlength{\tabcolsep}{6pt}
\renewcommand{\arraystretch}{1.15}
\small
\setlength{\aboverulesep}{1.0pt}
\setlength{\belowrulesep}{1.0pt}
\setlength{\cmidrulesep}{1.0pt}

\begin{subtable}[t]{0.9\textwidth}
\centering
\subcaption{Functional Role}
\label{tab:role-categories}
\begin{tabularx}{\textwidth}{
@{}Y
S[table-format=2.1, mode=text]
S[table-format=2.1, mode=text]
S[table-format=2.1, mode=text]
S[table-format=2.1, mode=text]@{}
}
\toprule
 & \multicolumn{2}{c}{\textbf{Human}} & \multicolumn{2}{c}{\textbf{LLM}} \\
\cmidrule(lr){2-3}\cmidrule(l){4-5}
\textbf{Category} & {Standard} & {Oracle} & {Standard} & {Oracle} \\
\midrule
Redirecting away from an unproductive path & 29.7 & 25.2 & 20.9 & 34.8 \\
Reframing the problem representation & 15.3 & 16.0 & 24.8 & 23.0 \\
Correcting local errors & 16.2 & 10.1 & 2.9 & 10.5 \\
Refocusing attention on salient evidence & 14.4 & 10.9 & 15.9 & 10.2 \\
Connecting partial insights into a coherent whole & 2.7 & 3.4 & 8.3 & 3.5 \\
Narrowing the search space & 4.5 & 1.7 & 7.1 & 3.1 \\
Prompting pattern recognition & 5.4 & 6.7 & 7.1 & 5.1 \\
Validating productive reasoning & 4.5 & 6.7 & 0.0 & 0.0 \\
Encouraging verification and self-checking & 1.8 & 5.9 & 0.0 & 2.0 \\
Extending a correct partial solution & 0.9 & 5.9 & 2.7 & 0.4 \\
Clarifying rules and constraints & 1.8 & 0.8 & 3.5 & 4.3 \\
Modeling a systematic strategy & 0.0 & 1.7 & 3.8 & 0.8 \\
Providing targeted hints or candidate answers & 0.9 & 3.4 & 1.8 & 2.0 \\
Providing closure or final confirmation & 0.9 & 1.7 & 0.0 & 0.0 \\
Explaining why an answer fits & 0.0 & 0.0 & 1.2 & 0.4 \\
Supporting persistence and reducing frustration & 0.9 & 0.0 & 0.0 & 0.0 \\
\bottomrule
\end{tabularx}
\end{subtable}

\vspace{10pt}

\begin{subtable}[t]{\textwidth}
\centering
\subcaption{Solution Leakage}
\label{tab:solution-categories}
\begin{tabularx}{0.9\textwidth}{
@{}Y
S[table-format=2.1, mode=text]
S[table-format=2.1, mode=text]
S[table-format=2.1, mode=text]
S[table-format=2.1, mode=text]@{}
}
\toprule
 & \multicolumn{2}{c}{\textbf{Human}} & \multicolumn{2}{c}{\textbf{LLM}} \\
\cmidrule(lr){2-3}\cmidrule(l){4-5}
\textbf{Category} & {Standard} & {Oracle} & {Standard} & {Oracle} \\
\midrule
Full solution revealed & 5.4 & 6.7 & 14.2 & 10.2 \tabularnewline
Strong narrowing or near-solution scaffold & 29.7 & 14.3 & 45.1 & 51.2 \tabularnewline
Key mechanism disclosure & 27.0 & 18.5 & 31.0 & 29.3 \tabularnewline
Minor local correction & 9.0 & 6.7 & 1.5 & 2.3 \tabularnewline
Answer revealed without full explanation & 5.4 & 5.0 & 5.3 & 5.1 \tabularnewline
Partial worked solution & 0.0 & 0.0 & 1.5 & 1.2 \tabularnewline
General reframing hint & 5.4 & 12.6 & 0.9 & 0.0 \tabularnewline
Wrong-path rejection & 7.2 & 11.8 & 0.0 & 0.0 \tabularnewline
Validation only & 5.4 & 10.9 & 0.0 & 0.0 \tabularnewline
Broad process hint & 1.8 & 5.9 & 0.3 & 0.0 \tabularnewline
No substantive reveal & 1.8 & 5.0 & 0.3 & 0.4 \tabularnewline
Method-level hint & 1.8 & 2.5 & 0.0 & 0.4 \tabularnewline
\bottomrule
\end{tabularx}
\end{subtable}

\vspace{-4pt}
\end{table*}

\subsection{Do interventions help students generalize to new problems (RQ3)?}
\label{sec:rq3}

\paragraph{Interventions do not consistently improve student generalization ability.} 
Across all three domains, neither with-context condition yielded a significant accuracy gain (\autoref{fig:generalization}). 
The \textit{Problem-Context} alone produced small domain-level shifts (math: $G=-0.01$; code: $G=-0.03$; brain: $G=+0.03$).
Similarly, the \textit{Intervention-Context} condition also did not improve generalization across domains (math: $G=-0.02$; code: $G=-0.04$; brain: $G=+0.04$), where each score $G$ is averaged over the four teacher models.
This indicates that access to teacher interventions and revised reasoning does not reliably translate into transferable problem-solving ability on new instances. 
One possible explanation is that interventions are overly specific to the original problem and emphasize instance-level corrections rather than more general, transferable strategies. 
As a result, the added context may fail to provide reusable insights, and in some cases, can even reduce performance when irrelevant information distracts the model \citep{wu2024easily, shi2023large}.

\subsection{How do LLMs compare to human behavior? (RQ4)}
\label{sec:rq4}

\paragraph{Humans intervene much less and much later than LLMs.}

In the \textit{Standard} monitoring condition, humans intervened less often and later than LLMs (Human: $\phi=0.74$, $\tau_{\text{rel}}=0.74$; LLM: $\phi=0.94$, $\tau_{\text{rel}}=0.24$; see \autoref{fig:human-llm}).
Human intervention behavior changed little across conditions ($\Delta\phi=+0.05$, $\Delta\tau_{\text{rel}}=-0.11$), while LLMs shifted their strategy substantially ($\Delta\phi=-0.23$, $\Delta\tau_{\text{rel}}=+0.33$).
To quantify how much human and LLM strategies diverged across conditions, we fit two Bayesian mixed models.\footnote{We fit a logistic mixed-effect model for intervention frequency and a Gaussian linear mixed-effect model for relative timing.
Both include fixed effects for agent (human/LLM), condition (Standard/Oracle), and their interaction, with a random intercept for question:
$Y_{ij} \sim \beta_0 + \beta_1\,\text{agent}_i + \beta_2\,\text{condition}_j + \beta_3\,(\text{agent} \times \text{condition})_{ij} + u_j,\; u_j \sim \mathcal{N}(0,\sigma^2_q)$.} 
We found that both outcomes showed a large interaction: frequency ($\beta=-2.60$, 95\% credible interval (CrI) $[-3.40, -1.80]$) and relative timing ($\beta=0.45$, 95\% CrI $[0.36, 0.53]$).
This interaction indicates that LLMs and humans respond to the kind of information available in different ways, such that while LLMs became significantly more selective and waited longer to intervene when they had full context, humans did not exhibit these changes.

Furthermore, on questions that the student would have solved correctly without intervention (6 out of 30 brain teasers), LLMs reduced their intervention rate when given full information (\textit{Standard}: $\phi_{\text{correct}}=0.72$; \textit{Oracle}: $\phi_{\text{correct}}=0.00$), whereas humans intervened at a similar rate across conditions (\textit{Standard}: $\phi_{\text{correct}}=0.37$; \textit{Oracle}: $\phi_{\text{correct}}=0.40$).
In these cases, rather than remaining silent, people sometimes praised or confirmed the student's answer (4 of 12 interventions), or they intervened to flag a flawed intermediate step in the reasoning, despite the student eventually arriving at the correct answer.
This suggests that people draw on richer pedagogical cues beyond answer correctness when deciding how to help.

To analyze the qualitative differences between human and LLM interventions, we categorized all messages (for the 30 selected brain teaser questions) along two dimensions: (i) \emph{functional role}, how the intervention supports the student, and (ii) \emph{solution leakage}, the extent to which the message directly discloses the solution (\autoref{tab:intervention-categories}; see \appref{Categories} for details).

\paragraph{Both LLMs and humans primarily intervene to redirect unproductive reasoning.} 

Across both monitoring conditions, human and LLM-based interventions most frequently redirected the student away from unproductive reasoning paths (\autoref{tab:role-categories}).
While LLMs also often reframed the problem representation ($\sim$24\%), humans more often relied on other strategies, such as correcting local errors ($\sim$13\%) and refocusing student attention on salient evidence ($\sim$13\%).
Finally, humans sometimes validated the student's reasoning process (up to 6.7\%), whereas the models never did.

\paragraph{LLMs often reveal substantial solution content, while humans offer more indirect guidance.} 

Overall, LLMs disclosed substantially more solution information than humans (\autoref{tab:solution-categories}). 
LLMs revealed the full solution roughly twice as often as humans (14.2\% vs.\ 
5.4\% in the \textit{Standard} condition, 10.2\% vs. 6.7\% in the \textit{Oracle} condition).
Models also heavily preferred providing near-complete solution scaffolding ($\sim$48\%), more so than humans did ($\sim$22\%).
Compared to models, humans instead relied more on wrong-path rejections and general reframing hints.

\section{Discussion}

In this work, we formalize helping as a sequential intervention game and introduce \textsc{Int-Bench}, a simulation-based benchmark for studying when, how, and how often LLMs intervene during problem solving. 
Using formal metrics and a complementary human study, we find that LLMs intervene more frequently and earlier than humans, often providing overly informative, instance-specific feedback that includes substantial solution content. 
% Furthermore, we find that teacher models are substantially more selective when given full information, suggesting that over-intervention can stem partly from partial observability and partly from a default propensity toward eagerness to help.
In contrast, in both partial and full information settings, humans tend to rely more on hints, scaffolding, and trajectory-aware guidance that leaves room for continued reasoning. 
These results highlight a tension between short-term task success and longer-term learning.
Though current AI assistants may be effective at correcting individual instances, they may be less reliable at preserving reasoning opportunities or supporting transfer to new problems.

AI assistants should ideally do more than correct an answer for a specific problem;
they should help users remain engaged, build the persistence needed to tackle future tasks independently, and become better general problem solvers. 
In learning settings, effort and productive struggle are often key ingredients to forming good understanding \citep{norton2012ikea, inzlicht2018effort, marsh2022influence}. 
More broadly, repeated exposure to overly direct assistance may gradually shift reasoning processes from humans to AI systems---a form of ``gradual disempowerment'' in which users become increasingly dependent on AI for how to reason through problems \citep{kulveit2025position, sharma2026whos}.
Recent work exploring assistants that are trained to maximize user empowerment \citep{ellis2025trainingllmagentsempower} indicates one potential direction, though designing assistants that balance short-term help with long-term success remains an open challenge.

Our experiments currently rely on LLM-simulated students.
Though this allows for more controlled, large-scale evaluation of intervention behavior, it is unclear how closely simulated student responses reflect the cognitive and motivational processes at play in human learning (e.g., misconceptions, cognitive load, and affective responses to feedback).
Additionally, our measure of generalization is limited to immediate transfer on a single related problem; 
however, interventions may influence learning over longer timescales, such as through repeated exposure and practice, which the current design does not capture.
Finally, extending \textsc{Int-Bench} to more student models, longer multi-episode interaction sequences, and real human learners would enable a more comprehensive account of AI interventions and their effects on users.

\section*{Acknowledgments}
We would like to thank the Supervised Program for Alignment Research (SPAR) for facilitating collaboration.
MKW was supported by Toyota Research Institute (TRI), Cooperative AI Foundation, the Foresight Institute, the Sony Research Award Program, UW-Tsukuba Amazon NVIDIA Cross Pacific AI Initiative, Jacobs CIFAR Research Fellowship, Templeton World Charity Foundation (https://doi.org/10.54224/34843).
TG was supported by grants from the Stanford Institute for Human-Centered Artificial Intelligence (HAI), Toyota Research Institute (TRI), and the Cooperative AI Foundation.

\bibliographystyle{plainnat}
\bibliography{references}

%%%%%%%%%%%%%%%%%%%%%%%%%%%%%%%%%%%%%%%%%%%%%%%%%%%%%%%%%%%%

\appendix

\addtocontents{toc}{\protect\setcounter{tocdepth}{2}}

\appendix

\setcounter{tocdepth}{2}
{\Large\textbf{Appendix}}
\tableofcontents
\newpage

% Custom tcolorbox style for compact prompts
\tcbset{
  compactbox/.style={
    coltitle=black,
    colframe=lightgray,
    colback=white,
    colbacktitle=white,
    boxrule=0.5pt,
    arc=2pt,
    boxsep=1pt,
    left=3pt,
    right=3pt,
    top=3pt,
    bottom=3pt,
    toptitle=1.5pt,
    bottomtitle=1.5pt,
    fonttitle=\bfseries\footnotesize,
    breakable
  }
}

% Global listings configuration for clean prompt formatting
\lstset{
  breakindent=0pt,
  breakautoindent=false
}

\section{Data}
\label{sec:data}

\subsection{Brain Teaser}
\label{braingle-data}
For the brain teaser dataset, we scraped brain teasers from the website Braingle.com, from the following categories: ``Language,'' ``Group,'' ``Rebus,'' and ``Riddle.'' 
Language teasers are those that involve the English language, often requiring one to think about and manipulate words and letters.
Group teasers involve recognizing groups and common attributes amongst words or letters.
Rebus brain teasers involve putting words or letters in interesting orientations to represent common phrases. 
Riddles are short poems or stories that describe something in a mysterious or indirect way.
They often can rhyme and pose a question with a hidden meaning.

To ensure the questions were suitable for our text-based models and human participants, we applied the following filters during scraping:
\begin{itemize}
    \item \textbf{Difficulty}: We restricted scraping to questions with a difficulty rating of $\le 2.0$ (out of $4.0$) to focus on questions that are solvable without too much time or specialized knowledge.
    \item \textbf{Popularity}: We required a popularity (``fun'') rating of $\ge 2.0$ (out of $4.0$) to ensure the questions were generally well-regarded and engaging.
\end{itemize}

After scraping, we further cleaned the dataset by removing any questions containing images and keywords suggesting visuals. 
From the remaining pool, we aggregated a dataset of 500 questions by random sampling, after ensuring the inclusion of our manually selected subset.

For the human study, we manually selected a subset of 30 questions. 
The selection process involved reviewing the top 500 scraped questions (ranked by most popular and least difficult) and choosing those that did not involve multiple subparts, require extensive external or domain knowledge, or contain overly difficult vocabulary.
These 30 questions were used in both the model simulations and for the human experiment.

\subsection{Simulation Data}

Our simulation consisted of 18,000 episodes per monitoring condition (500 questions $\times$ 3 domains $\times$ 4 teacher models $\times$ 3 runs), for a total of 36,000 teacher rollouts across the \textit{Standard} and \textit{Oracle} conditions.
The \textit{Standard-Continue} and \textit{Stop-and-Answer} post-intervention reasoning conditions included only the \textit{Standard} monitoring condition episodes where the \textit{Standard} teacher chose to intervene---16,111 and 16,110 episodes, respectively, aggregated across all three domains (small differences between the two reflect miscellaneous API errors).
\textit{Oracle-Continue} was further restricted to \textit{Oracle} teacher interventions with a valid recoverable timing anchor.

% \vspace{0.3em}
\begin{table}[b]
\caption{Number of episodes used in each post-intervention analysis condition, per domain. 
}
\centering
\begin{tabular}{lrrrr}
\toprule
Condition & Code Debugging & Mathematics & Brain Teasers & Total \\
\midrule
Standard-Continue  & 5{,}705 & 4{,}727 & 5{,}679 & 16{,}111 \\
Stop-and-Answer    & 5{,}704 & 4{,}727 & 5{,}679 & 16{,}110 \\
Oracle-Continue    & 2{,}128 & 1{,}236 & 4{,}082 &  7{,}446 \\
\bottomrule
\end{tabular}
\label{tab:episode-counts}
\end{table}

% \vspace{0.3cm}

\paragraph{Oracle Timing Recovery}
\label{sec:timing-recovery}

In the \textit{Oracle} monitoring condition, the teacher reports its chosen intervention point as a 5-word verbatim quote from the reasoning transcript.
We recover the character position by searching for this quote using a three-stage procedure: (1) exact substring match, (2) case-insensitive match, and (3) flexible-whitespace token regex (preserving operators and punctuation).
Approximately 23\% of \textit{Oracle} interventions were excluded (2{,}241 of 9{,}695 intervened episodes): $\sim$16\% (1{,}529 episodes) because the quote could not be located, $\sim$5\% (477 episodes) because the recovered position fell in the final answer region rather than the reasoning trace, $\sim$2\% (229 episodes) due to multiple possible matches, and a small remainder (6 episodes) in which the teacher response contained no quote at all.
Retention rates varied by domain: 83.0\% for brain teasers, 70.2\% for code debugging, and 71.5\% for mathematics.
The average retained $\sim$77\% (7{,}446 episodes) formed the basis of all \textit{Oracle} teacher timing statistics ($\tau_{\text{rel}}$, $\tau_{\text{abs}}$) and \textit{Oracle-Continue} episodes.

\clearpage
\section{Human Study}
\label{sec:human_study}

\begin{figure}[H]
  \centering
  \includegraphics[width=0.95\linewidth]{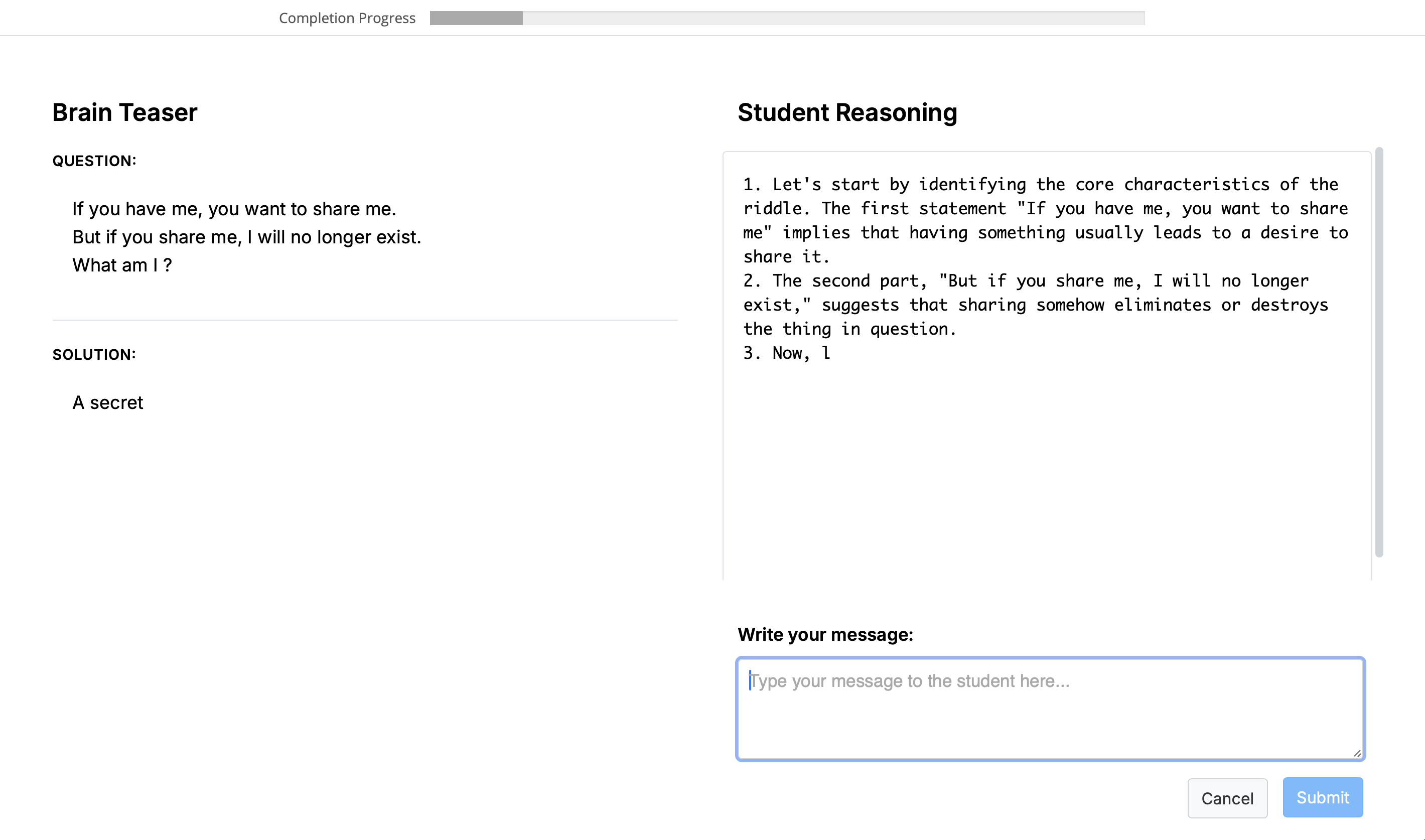}
  \caption{Interface for the human study. Participants see the brain teaser information (question and solution) on the left side of the screen. The right side of the screen shows the student's reasoning trace, which can be revealed by pressing the right arrow key. Participants can pause and intervene by pressing the space bar, at which point a text box appears where they can type their intervention message.}
  \label{fig:Interface}
\end{figure}

% \subsection{Power Analysis}

% We conducted a power analysis to ensure our sample sizes were sufficient to detect meaningful effect sizes, given the 30 brain teaser questions used in the study.

% We evaluated the difference between humans and LLMs using paired tests on the per-question means. 
% For comparing intervention frequency between humans and LLMs in the \textit{Standard} condition, we observed a large effect size (Cohen's $d = 1.157$). 
% With 30 questions, our study achieved nearly 100\% power to detect this effect at a $\alpha = 0.05$ significance level. 
% Similarly, for the relative timing of interventions, the observed difference in the \textit{Standard} condition yielded a very large effect size (Cohen's $d = 3.135$), resulting in nearly 100\% statistical power.

% In the \textit{Oracle} condition, the difference in intervention frequency between humans and LLMs resulted in a smaller effect size (Cohen's $d = 0.300$), achieving 37.6\% power. 
% However, the relative timing of interventions in the \textit{Oracle} condition still yielded a large effect size (Cohen's $d = 1.871$), achieving nearly 100\% statistical power (based on the 24 questions where both humans and LLMs intervened).

\subsection{Participants}
We recruited participants via the crowd-sourcing platform Prolific.
We included participants who are fluent in English, based in the United States, have an approval rate of at least 98\%, and have completed at least 300 previous studies.
Participants were compensated at a rate of \$12 per hour.
All studies were approved by our Institutional Review Board (IRB).

\subsection{Procedure}
Participants first read instructions explaining the experimental setup. 
They were asked to imagine that they are a teacher working with some students who are trying to solve brain teasers. 
They were told that their goal as the teacher is to help the students solve the problems.
After reading the instructions, participants completed two example trials and a comprehension check before proceeding to the main trials.
For each trial, only after acknowledging that they had read the question and understood the solution could they begin revealing the student's reasoning trace.

At the end of the study, participants were asked how many years of teaching experience they have, whether they are a parent or guardian of children (if so, their age ranges), whether they used AI tools during the experiment (if so, how they used it), how they decided whether and when to intervene, and any additional feedback they might have.

\subsection{Participant Demographics Analysis}
\label{sec:human-demog-analysis}
In the \textit{Standard} condition, 52\% of participants had no teaching experience, 24\% had less than 1 year, and 24\% had over 1 year of teaching experience. 48\% were parents or guardians of children.
In the \textit{Oracle} condition, 60\% had no teaching experience, 12\% had less than 1 year, and 28\% had over 1 year of teaching experience. 68\% were parents or guardians of children.

We examined whether teaching experience and parenthood status affected intervention behavior.
For each, we split participants into two groups (no teaching experience vs. teaching experience, and non-parent vs. parent) and compared per-participant means on intervention frequency ($\phi$) and relative timing ($\tau_{\text{rel}}$) using two-sided Mann-Whitney U tests.

\paragraph{Teaching Experience}
We split participants into those with no teaching experience (\textit{no experience}: Standard $n=13$, Oracle $n=15$) and those with any teaching experience (\textit{experienced}: Standard $n=12$, Oracle $n=10$).
Pooling both monitoring conditions, experienced and no-experience participants showed similar intervention rates and timing ($\phi$: $U=293.0$, $p=0.769$; $\tau_{\text{rel}}$: $U=213.0$, $p=0.093$).

\begin{figure}[H]
    \centering
    \includegraphics[width=0.85\linewidth]{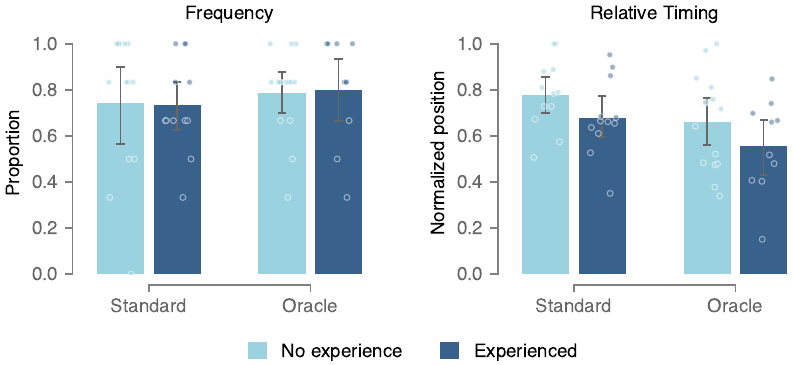}
    \caption{Intervention frequency ($\phi$, left) and relative timing ($\tau_{\text{rel}}$, right) for no-experience vs.\ experienced participants, shown for Standard and Oracle monitoring conditions. Error bars represent 95\% bootstrapped CIs. Participants are indicated by individual points.}
    \label{fig:experience-comparison}
\end{figure}

\paragraph{Parenthood}
For parenthood, we split participants into parents (or guardians) (\textit{parent}: Standard $n=12$, Oracle $n=17$) and non-parents (\textit{non-parent}: Standard $n=13$, Oracle $n=8$).
Neither intervention frequency nor relative timing differed between groups ($\phi$: $U=366.0$, $p=0.213$; $\tau_{\text{rel}}$: $U=318.0$, $p=0.635$).

Overall, these results suggest that neither teaching experience nor parenthood reliably modulates intervention frequency or timing in our paradigm, supporting the robustness of the human baseline across participant backgrounds.

\begin{figure}[H]
    \centering
    \includegraphics[width=0.85\linewidth]{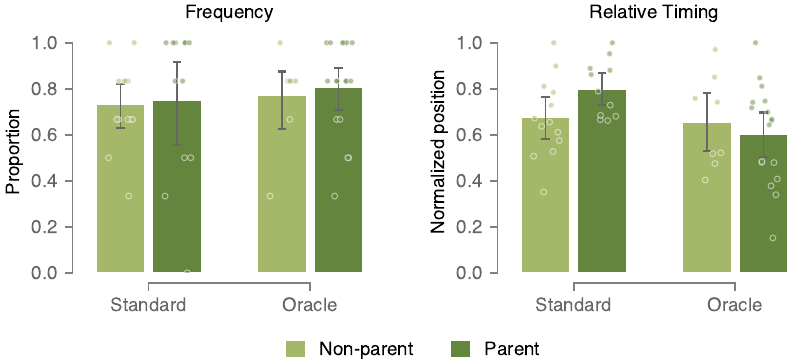}
    \caption{Intervention frequency ($\phi$, left) and relative timing ($\tau_{\text{rel}}$, right) for non-parent vs.\ parent participants, shown for Standard and Oracle monitoring conditions. Error bars represent 95\% bootstrapped CIs. Participants are indicated by individual points.}
    \label{fig:parenthood-comparison}
\end{figure}

\clearpage
\section{Pipeline Details}

% \subsection{Models}
% With the exception of GPT-5.2 and Gemini 3 Flash, we used the OpenRoute API to access models. 
% We used the OpenAI API for GPT-5.2, and Google API for Gemini 3 Flash.

\subsection{Math Evaluation}
For the MATH-500 dataset, we used a deterministic grading methodology rather than an LLM-based judge \citep{lightman2023letsverifystepstep}.
The grading process includes two stages: (1) normalizing the student's answer and the ground truth to a canonical format, and (2) checking for mathematical equivalence using SymPy. 
This ensures that our evaluation is objective and reproducible, avoiding the potential variability and biases of LLM-based grading for math questions.

\subsection{Problem Generation}
\label{sec:problem-generation}

\autoref{tab:problem-generation} summarizes the functionality and role of each component in our structured problem-generation pipeline.
In our setup, we generated 5 candidate variants per reference problem. 
We used GPT-5.2 as the model for all parts of the pipeline.
Prompts for each component are provided in \secref{sec:prompts-generation}.

\begin{table*}[t]
\caption{Specifications of the structured problem-generation pipeline. The skill extractor $\mathcal{E}$ identifies latent reasoning skills, the clusterer~$\mathcal{C}$ constructs a taxonomy of high-level categories of skills, the generator $\mathcal{G}$ produces skill-preserving problem variants, and the validator $\mathcal{V}$ enforces correctness of generated problems across domains.}
\centering
\small
\renewcommand{\arraystretch}{1.1}
\begin{tabularx}{\textwidth}{@{} >{\RaggedRight}p{2.75cm} >{\hsize=1.3\hsize}Y >{\hsize=0.85\hsize}Y >{\hsize=0.85\hsize}Y @{}}
\toprule
Component & Description & Inputs & Outputs \\ 
\midrule
Skill Extractor $\mathcal{E}$ &
Identifies the latent reasoning skill or capability required to solve a reference problem. &
Problem statement; and baseline solution  &
Fine-grained skill label $s_i$. \\
\addlinespace
Skill Clusterer $\mathcal{C}$ &
Groups similar fine-grained skills into high-level categories to construct a consistent dataset-level taxonomy. &
Set of extracted skills $\{s_i\}$ from skill extractor $\mathcal{E}$. &
High-level skill categories $\mathcal{S} = \{S_1, \dots, S_K\}$. \\
\addlinespace
Problem Generator $\mathcal{G}$ &
Generates candidate variants that preserve the target skill category while modifying surface form, context, or parameters. &
Reference problem $Q$; assigned skill category $S_k$; generation constraints; $N$. &
Candidate question set $\tilde{\mathcal{Q}} = \{\tilde{Q}_1, \dots, \tilde{Q}_N\}$ and candidate answer set $\tilde{\mathcal{Y}} = \{\tilde{Y}_1, \dots, \tilde{Y}_N\}$. \\
\addlinespace
Problem Validator $\mathcal{V}$ &
Verifies correctness of generated variants and filters out invalid or misaligned problems. &
Candidate variants $(\tilde{\mathcal{Q}}, \tilde{\mathcal{Y}})$; skill categories $\mathcal{S}$. &
Validated set $(\tilde{\mathcal{Q}}, \tilde{\mathcal{Y}})_{\text{valid}} \subseteq (\tilde{\mathcal{Q}}, \tilde{\mathcal{Y}})$. \\
\bottomrule
\end{tabularx}
\label{tab:problem-generation}
\end{table*}

\subsection{Categorization of Interventions}
\label{Categories}
We developed a bottom-up categorization pipeline to label both human and LLM interventions along two complementary dimensions: (i) the \emph{functional role} of the intervention in supporting the student, and (ii) the \emph{degree of solution information revealed}. 

Let $\mathcal{M}$ denote the set of all intervention messages. 
Our pipeline proceeds in three stages. 
First, each intervention message $m \in \mathcal{M}$ is passed to an LLM $L_1$, which assigns an initial high-level category $c \in \mathcal{C}$.
Second, the set of all generated categories $\mathcal{C}$ is aggregated and provided to a separate LLM $L_2$, which clusters semantically similar categories to construct a consolidated taxonomy $\mathcal{C}^\star$. 
Finally, each message $m$ is relabeled by a third LLM $L_3$, which assigns a final category $\hat{c} \in \mathcal{C}^\star$, ensuring consistency and comparability across the dataset. 
We used GPT-5.2 as the model for each stage of this pipeline.
All prompts are provided in \secref{sec:prompts-categorization}.

\clearpage
\section{Results}
\label{sec:results}

\vspace{-1cm}

\subsection{LLM Interventions}

\begin{figure}[h!]
    \centering
    \includegraphics[width=0.8\linewidth]{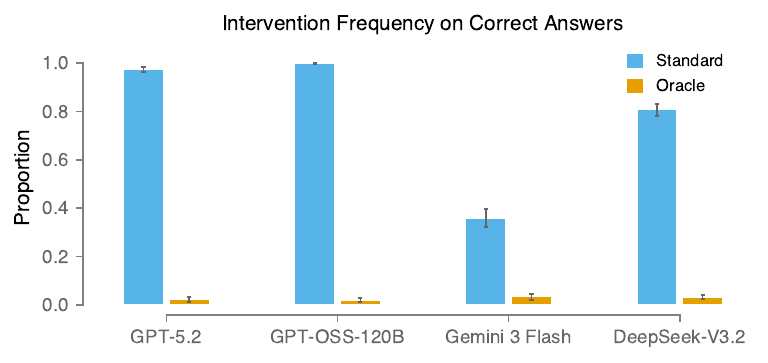}
    \caption{Frequency of interventions on problems where the student's unassisted baseline answer was correct ($\phi_{\text{correct}}$), shown across different teacher models and evaluation domains. Error bars indicate 95\% bootstrapped confidence intervals.}
    \label{fig:phi_correct_appendix}
\end{figure}

\begin{figure}[h]
    \centering
    \includegraphics[width=0.8\linewidth]{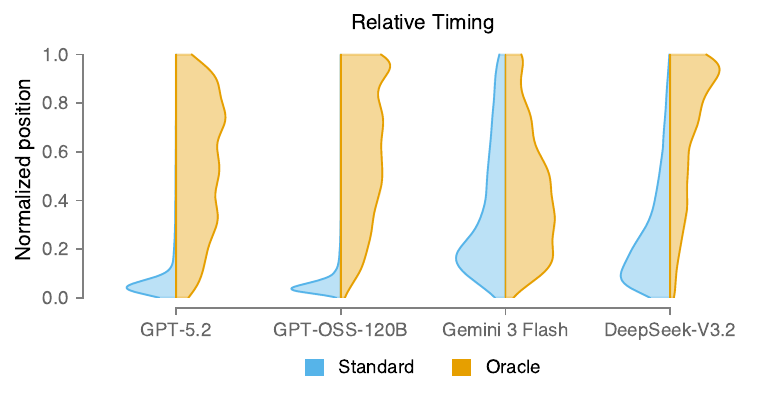}
    \caption{Relative intervention timing ($\tau_{\text{rel}}$) across all models and domains for both \textit{Standard} and \textit{Oracle} conditions. 
    In the \textit{Standard} condition, models intervene much earlier in the reasoning process compared to the \textit{Oracle} condition, where they have access to the full trace, student answer, and correctness, and can select a more optimal point to intervene.}
    \label{fig:relative_timing_appendix}
\end{figure}

\begin{table}[H]

\caption{Intervention metrics (intervention frequency $\phi$, frequency on correct answers $\phi_{\text{correct}}$, relative timing $\tau_{\text{rel}}$, and absolute character-based timing $\tau_{\text{abs}}$) across all teacher models, domains, and monitoring conditions.}

\centering
% \small
\begin{tabular}{l|cccc|cccc}
\toprule
& \multicolumn{4}{c|}{\textbf{Standard}} & \multicolumn{4}{c}{\textbf{Oracle}} \\
\textbf{Model} & $\phi$ & $\phi_{\text{correct}}$ & $\tau_{\text{rel}}$ & $\tau_{\text{abs}}$ & $\phi$ & $\phi_{\text{correct}}$ & $\tau_{\text{rel}}$ & $\tau_{\text{abs}}$ \\
\midrule
\multicolumn{9}{c}{\textit{Code Debugging}} \\
\midrule
GPT-5.2 & 1.00 & 1.00 & 0.04 & 51.67 & 0.57 & 0.06 & 0.55 & 1033.66 \\
GPT-OSS-120B & 1.00 & 1.00 & 0.05 & 58.17 & 0.41 & 0.02 & 0.70 & 1491.13 \\
Gemini 3 Flash & 0.83 & 0.74 & 0.35 & 499.60 & 0.49 & 0.06 & 0.48 & 760.93 \\
DeepSeek-V3.2 & 0.98 & 0.96 & 0.18 & 243.56 & 0.56 & 0.07 & 0.60 & 1179.36 \\
\midrule
\multicolumn{9}{c}{\textit{Mathematics}} \\
\midrule
GPT-5.2 & 0.97 & 0.95 & 0.15 & 150.21 & 0.29 & 0.01 & 0.57 & 1047.76 \\
GPT-OSS-120B & 1.00 & 1.00 & 0.07 & 67.03 & 0.30 & 0.02 & 0.58 & 1096.61 \\
Gemini 3 Flash & 0.37 & 0.13 & 0.46 & 760.89 & 0.27 & 0.01 & 0.48 & 860.10 \\
DeepSeek-V3.2 & 0.81 & 0.74 & 0.33 & 401.81 & 0.30 & 0.01 & 0.77 & 1468.33 \\
\midrule
\multicolumn{9}{c}{\textit{Brain Teaser}} \\
\midrule
GPT-5.2 & 1.00 & 1.00 & 0.05 & 59.55 & 0.86 & 0.00 & 0.51 & 621.17 \\
GPT-OSS-120B & 1.00 & 1.00 & 0.06 & 61.86 & 0.73 & 0.02 & 0.56 & 690.85 \\
Gemini 3 Flash & 0.86 & 0.26 & 0.36 & 423.75 & 0.85 & 0.04 & 0.42 & 515.07 \\
DeepSeek-V3.2 & 0.93 & 0.68 & 0.28 & 319.42 & 0.84 & 0.04 & 0.71 & 893.48 \\
\bottomrule
\end{tabular}

\label{tab:intervention_results_detailed}
\end{table}

\vspace{1cm}

% \clearpage
\subsection{Human vs. LLM Brain Teaser Comparisons}

We compare human vs. LLM intervention behavior using the 30 manually selected brain teaser questions described in \secref{braingle-data}. 
On the human side, we collected 150 human episodes per condition (25 participants $\times$ 6 episodes each, i.e., each question has 5 human annotations) and 360 LLM episodes per condition (30 questions $\times$ 4 teacher models $\times$ 3 repetitions). 

Humans in the \textit{Standard} condition intervened on 111/150 episodes, and 119/150 in the \textit{Oracle} condition.
LLMs in the \textit{Standard} condition intervened on 339/360 episodes.
In the \textit{Oracle} condition, they attempted to intervene on 256/360 episodes. 
However, 31 of the 256 interventions from the \textit{Oracle} LLMs could not be assigned a character position by the recovery procedure in \secref{sec:timing-recovery}. 
Timing-based results in this section therefore used $n=225$ LLM \textit{Oracle} interventions (frequency and category-based statistics are unaffected).

\begin{figure}[h]
    \centering
    \includegraphics[width=0.95\linewidth]{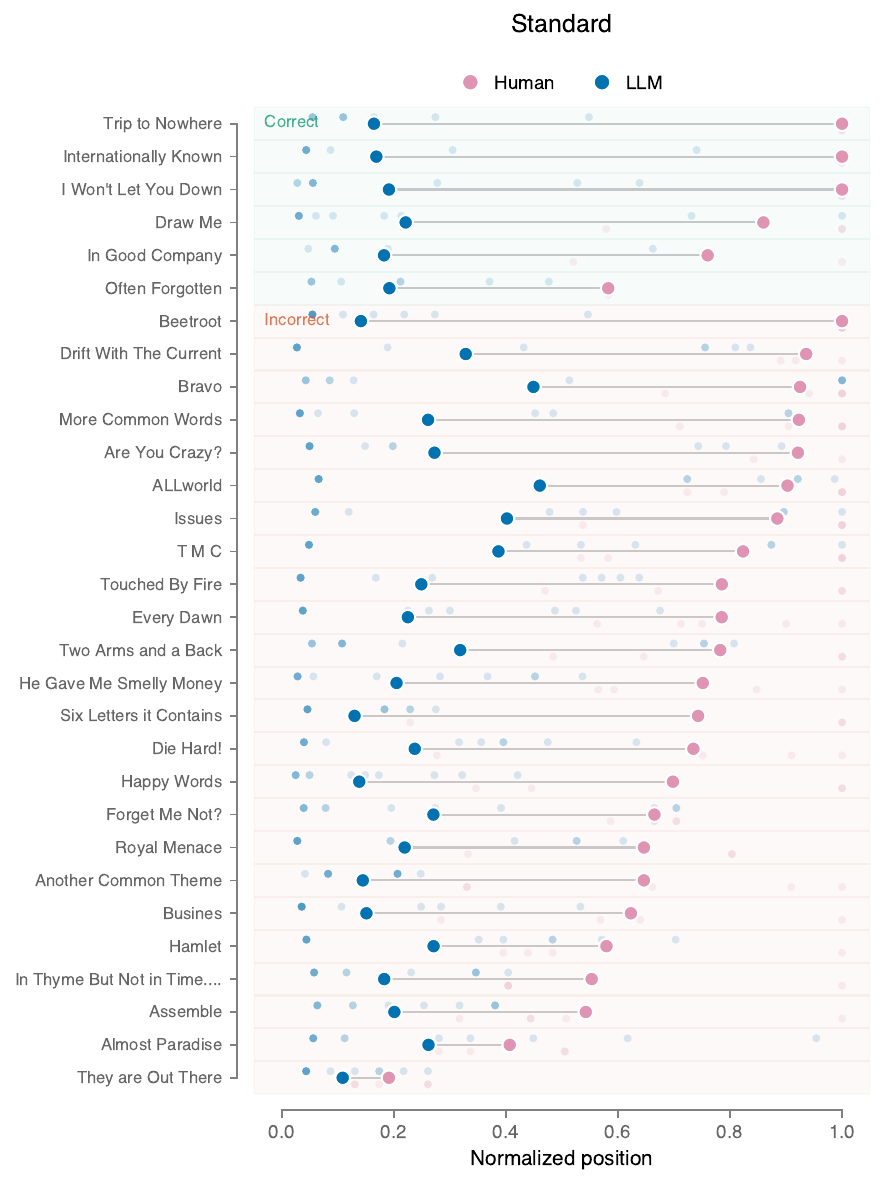}
    \caption{Trial comparison of mean intervention timing between human and LLM teachers in the \textit{Standard} condition across the 30 brain teaser questions used in the human study. LLM results are averaged across all four evaluated models. Questions are grouped by whether the student's baseline initial answer (i.e., prior to any intervention) was correct or incorrect. Small dots represent individual episodes.}
    \label{fig:human_llm_standard_timing_trials}
\end{figure}

\begin{figure}
    \centering
    \includegraphics[width=0.95\linewidth]{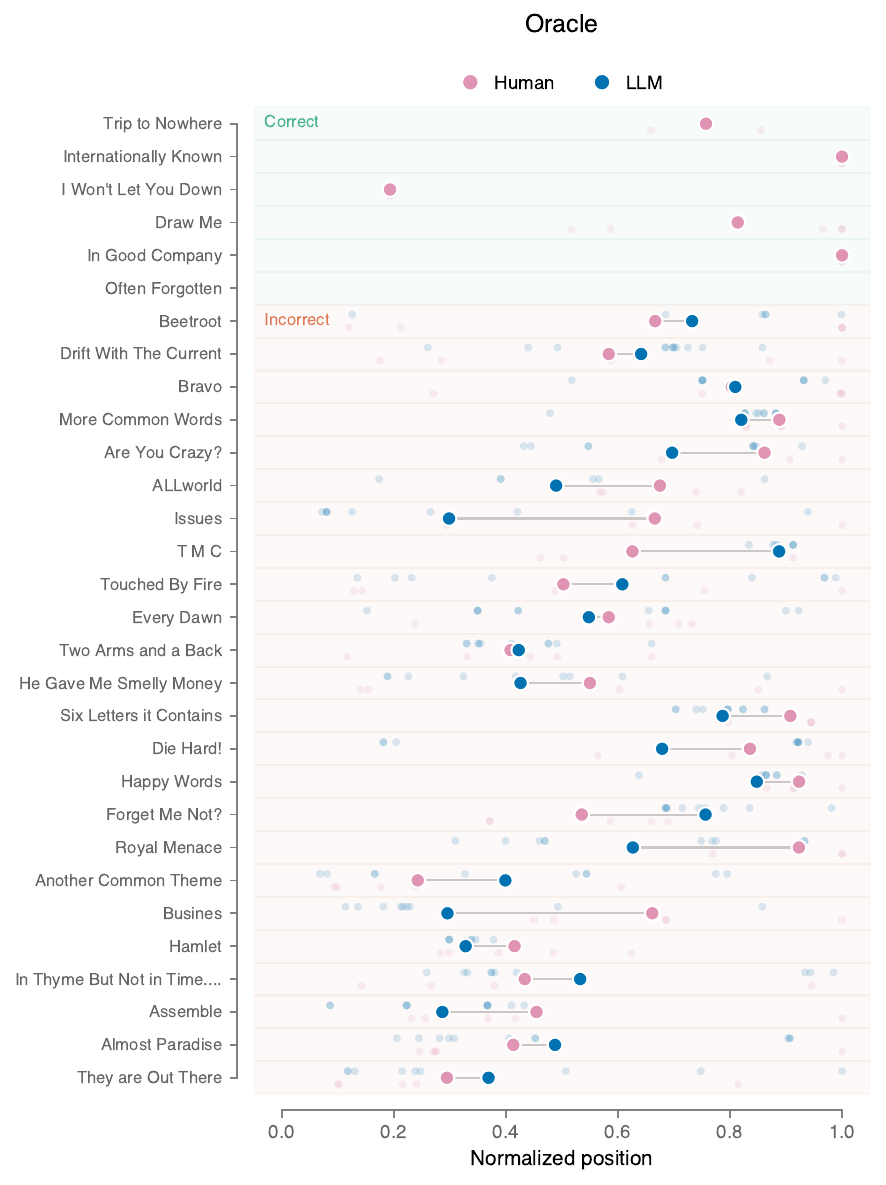}
    \caption{Trial comparison of mean intervention timing between human and LLM teachers in the \textit{Oracle} condition across the 30 brain teaser questions used in the human study. LLM results are averaged across all four evaluated models. Questions are grouped by whether the student's baseline initial answer (i.e., prior to any intervention) was correct or incorrect. Small dots represent individual episodes.}
    \label{fig:human_llm_oracle_timing_trials}
\end{figure}

\clearpage
\subsection{Examples}

% \paragraph{Human}

\begin{table*}[h]
    \caption{Example of a brain teaser and human intervention collected during the study.}
    \centering
    \small
    \begin{tabular}{p{0.22\linewidth} p{0.73\linewidth}}
    \toprule
    \textbf{Question} & \textbf{Title}: In Thyme But Not in Time... \newline 
    \textbf{Content}: My first is in riddle, but not in little.
                    My second is in think, but not in brink.
                    My third is in thyme, but not in time.
                    My fourth is in mother, but not in brother.
                    My last is in time, but not in climb.
                    What am I?
                      \newline 
    \textbf{Solution}: The word rhyme. \\
    \midrule
    \textbf{Student Reasoning} & 1. First letter: "My first is in riddle, but not in little."
        - "Riddle" contains 'r', but "little" does not contain 'r'.
        - Therefore, the first letter is 'r'.

      2. Second letter: "My second is in think, but not in brink."
        - "Think" contains 'i', but "brink" does not contain 'i'.
        - Therefore, the second letter is 'i'.

      3. Third letter: "M
      \\
    \midrule
    \textbf{Intervention} & Brink contains an i. Try again \\
    \bottomrule
    \end{tabular}
    \label{tab:human_study_example1}
\end{table*}

\vspace{1cm}

\begin{table*}[h]
    \caption{Example of a brain teaser and human intervention collected during the study.}
    \centering
    \small
    \begin{tabular}{p{0.22\linewidth} p{0.73\linewidth}}
    \toprule
    \textbf{Question} & \textbf{Title}: Beetroot \newline 
    \textbf{Content}: What advice is shown below?
                      BEETROOT
                      YEETROOT
                      YOETROOT
                      YOUTROOT
                      YOURROOT
                      YOURSOOT
                      YOURSEOT
                      YOURSELT
                      YOURSELF
                      \newline 
    \textbf{Solution}: Be true to yourself. \newline
    BEETROOT to YOURSELF (sound it out) \\
    \midrule
    \textbf{Student Reasoning} & 1. The given sequence of words starts with "BEETROOT" and each subsequent line has one letter moved from the end to the beginning. \newline
      2. We need to observe the pattern in the rearrangement process:
        - BEETROOT -> YEETROOT (Y moves from last to first)
        - YEETROOT -> YOETROOT (O moves from last to first)
        - YOETROOT -> YOUTROOT (U moves from last to first)
        - YOUTROOT -> YOURROOT (R moves from last to first)
        - YOURROOT -> YOURSOOT (S moves from last to first)
        - YOURSOOT -> YOURSEOT (E moves from last to first)
        - YOURSEOT -> YOURSELT (L moves from last to first)
        - YOURSELT -> YOURSELF (F moves from last to first) \newline
      3. Each word maintains its original structure while progressively repositioning the last letter until it reaches the front, forming the word "YOURSELF". \newline
      4. The process essentially shows how to spell out "YOURSELF" by moving the letters from the end to the beginning one at a time.
      \\
    \midrule
    \textbf{Intervention} & Why your answer is true, you need to simplfy.  Simply state the phrase outloud and you will hear the correct answer. \\
    \bottomrule
    \end{tabular}

    \label{tab:human_study_example2}
\end{table*}

\vspace{1cm}

% \clearpage
% \paragraph{LLM}

\begin{table*}[h]
    \caption{Example of an intervention from an LLM-based teacher (GPT-5.2) in the \textit{Standard} condition.}
    \centering
    \small
    \begin{tabular}{p{0.22\linewidth} p{0.73\linewidth}}
    \toprule
    \textbf{Question} & \textbf{Title}: Beetroot \newline 
    \textbf{Content}: What advice is shown below?
                      BEETROOT
                      YEETROOT
                      YOETROOT
                      YOUTROOT
                      YOURROOT
                      YOURSOOT
                      YOURSEOT
                      YOURSELT
                      YOURSELF
                      \newline 
    \textbf{Solution}: Be true to yourself. \newline
    BEETROOT to YOURSELF (sound it out) \\
    \midrule
    \textbf{Student Reasoning} & 1. The given sequence of words starts with "BEETRO
      \\
    \midrule
    \textbf{Intervention} & Try reading the first and last words aloud and “hearing” the phrase they form: BEETROOT → “be true to” and YOURSELF stays “yourself.” The intermediate steps just morph the letters to get from one to the other. \\
    \bottomrule
    \end{tabular}

    \label{tab:llm_example1}
\end{table*}

\clearpage

\section{Additional Experiments}
\label{sec:additional-experiments}

\subsection{Increment Sizes}
\label{sec:chunking}
We investigated the effect of increment size on intervention behavior.
We experimented with revealing the reasoning in 300-character and per-sentence increments. 
From \autoref{tab:chunking_metrics}, we observed that intervention frequency remained unchanged across all conditions and the relative position is consistently small. 
Moreover, the absolute intervention position ($\tau_{\text{abs}}$) closely tracks the increment size itself, indicating that the teacher intervened immediately after viewing the first increment, regardless of granularity.

\begin{table}[h]
\caption{Impact of different increment sizes on intervention frequency and timing in the code debugging domain, with GPT-5.2 as the teacher model in the \textit{Standard} condition.}
\centering
\begin{tabular}{lccc}
\toprule
\textbf{Increment Size} & $\phi$ & $\tau_{\text{rel}}$ & $\tau_{\text{abs}}$ \\
\midrule
50-character & 1.00 & 0.04 & 50.6 \\
300-character & 1.00 & 0.25 & 301.9 \\
1-sentence & 1.00 & 0.11 & 138.0 \\
\bottomrule
\end{tabular}

\label{tab:chunking_metrics}
\end{table}

\subsection{Student Models}
\label{sec:student-models}
We examined whether the student model affects teacher intervention behavior by varying the student across a larger model (Qwen3-32B), a different model family (Llama-3.1-8B), and the same model as the teacher (GPT-5.2). 
As shown in \autoref{tab:student_metrics}, intervention frequency and timing remained nearly identical across all student models.

\begin{table}[h]
\caption{Impact of different student models on teacher intervention frequency and timing in the code debugging domain, with GPT-5.2 as the teacher model in the \textit{Standard} condition.}
\centering
\begin{tabular}{lccc}
\toprule
\textbf{Student Model} & $\phi$ & $\tau_{\text{rel}}$ & $\tau_{\text{abs}}$ \\
\midrule
Qwen2.5-7B & 1.00 & 0.04 & 50.6 \\
GPT-5.2 & 1.00 & 0.07 & 51.0 \\
Llama-3.1-8B & 1.00 & 0.04 & 50.1 \\
Qwen3-32B & 1.00 & 0.08 & 50.3 \\
\bottomrule
\end{tabular}

\label{tab:student_metrics}
\end{table}

\subsection{Prompt Sensitivity}
\label{sec:prompt-sensitivity}
We conducted a prompt sensitivity analysis to verify that intervention behavior is robust to prompt wording. 
We tested two prompt variations: ``Variation 1'' adds an explicit constraint to only intervene if truly necessary, while ``Variation 2'' instructs the teacher to scaffold and guide the student. 
As shown in \autoref{tab:prompt_metrics}, the results are consistent across all prompt conditions: the \textit{Standard} teacher intervened at nearly the same rate and timing regardless of the prompt variation, and the \textit{Oracle} teacher also showed similar behavior across all settings.

\begin{table}[h]
\caption{Prompt sensitivity of intervention frequency and timing for both monitoring conditions in the code debugging domain with GPT-5.2.}
\centering
\begin{tabular}{l|ccc|ccc}
\toprule
 & \multicolumn{3}{c|}{\textbf{Standard}} & \multicolumn{3}{c}{\textbf{Oracle}} \\
% \cmidrule(lr){2-4}\cmidrule(lr){5-7}
\textbf{Prompt} & $\phi$ & $\tau_{\text{rel}}$ & $\tau_{\text{abs}}$ & $\phi$ & $\tau_{\text{rel}}$ & $\tau_{\text{abs}}$ \\
\midrule
Original & 1.00 & 0.04 & 50.5 & 0.53 & 0.53 & 1027.6 \\
Variation 1 & 1.00 & 0.04 & 51.5 & 0.49 & 0.55 & 1111.1 \\
Variation 2 & 1.00 & 0.04 & 50.4 & 0.51 & 0.53 & 1067.0 \\

% \centering
% \begin{tabular}{lccc}
% \toprule
% \textbf{Prompt} & $\phi$ & $\tau_{\text{rel}}$ & $\tau_{\text{abs}}$ \\
% \midrule
% \multicolumn{4}{c}{\textit{Standard}} \\
% \midrule
% Original    & 1.00 & 0.04 & 50.5 \\
% Variation 1 & 1.00 & 0.04 & 51.5 \\
% Variation 2 & 1.00 & 0.04 & 50.4 \\
% \midrule
% \multicolumn{4}{c}{\textit{Oracle}} \\
% \midrule
% Original    & 0.53 & 0.53 & 1027.6 \\
% Variation 1 & 0.49 & 0.55 & 1111.1 \\
% Variation 2 & 0.51 & 0.53 & 1067.0 \\

\bottomrule
\end{tabular}

\label{tab:prompt_metrics}
\end{table}

\clearpage
\section{Prompts}
\label{sec:prompts}

We provide the prompt templates used for our models below. For brevity, we show the templates for the brain teaser domain, as they are representative of the structure used across all domains.

\subsection{Baseline Student}
\label{sec:student-prompts}

The student model uses a chain-of-thought (CoT) prompt to generate its reasoning and final answer.

\begin{tcolorbox}[compactbox, title=Student Baseline Prompt, width=\textwidth]
\begin{lstlisting}[breaklines=true, basicstyle=\footnotesize\ttfamily, escapeinside={(*@}{@*)}]
(*@\textbf{[System]}@*)
Your task is to solve a brain teaser.
You must think step-by-step and show your full reasoning. Do not summarize it.

Your response must follow this format. Do not mix your reasoning with the final answer.

Reasoning:
<your step-by-step reasoning>

Final Answer: <answer>

(*@\textbf{[User]}@*)
Title: {{ question_title }}
Problem: {{ question_content }}
\end{lstlisting}
\end{tcolorbox}

The post-intervention and generalization student prompts share this system prompt (with minor domain-specific wording variation). 
Only the user message turns are shown in those subsections below.

\clearpage
\subsection{Teacher}
\label{sec:teacher-prompts}

\paragraph{Standard Monitoring Condition}
The \textit{Standard} teacher watches the student's reasoning in chunks and decides whether to intervene.

\begin{tcolorbox}[compactbox, title=Standard Teacher Prompt, width=\textwidth]
\begin{lstlisting}[breaklines=true, basicstyle=\footnotesize\ttfamily, escapeinside={(*@}{@*)}]
(*@\textbf{[System]}@*)
You are an expert teacher watching a student attempt a brain teaser. Your goal is to teach the student how to solve brain teasers effectively and efficiently.

You will be shown a brain teaser, portions of the student's reasoning transcript, and a reference solution. Only you as the teacher have access to the reference solution; the student does not.

You will see the reasoning in increments of {{ chunk_size }} characters at a time. After each increment, you will be asked whether you would like to intervene. If you choose not to intervene, you will then receive the next {{ chunk_size }} characters, along with all previously shown text. You may only intervene once. If you choose to intervene, the task ends immediately and your intervention message will be sent to the student. You may also choose not to intervene.

Your response must follow this format. Do not include any explanations or additional text.

Intervene: [Yes/No]

Intervention: <If 'Yes', write your intervention message to the student at this moment. If 'No', leave this blank.>


(*@\textbf{[User]}@*)
Title: {{ question_title }}
Problem: {{ question_content }}

Student reasoning so far: {{ transcript_portion }}

Reference solution: {{ reference_content }}
\end{lstlisting}
\end{tcolorbox}

\clearpage
\paragraph{Oracle Monitoring Condition}
The \textit{Oracle} teacher model has access to the full student transcript, final answer, and judge verdict before deciding whether and when to intervene.

\begin{tcolorbox}[compactbox, title=Oracle Teacher Prompt, width=\textwidth]
\begin{lstlisting}[breaklines=true, basicstyle=\footnotesize\ttfamily, escapeinside={(*@}{@*)}]
(*@\textbf{[System]}@*)
You are an expert teacher. Your goal is to teach the student how to solve brain teasers effectively.

You will be shown (1) a brain teaser, (2) the full transcript of the student's reasoning process, (3) the student's final answer, (4) a reference solution, and (5) whether the student's answer is correct/incorrect. Only you as the teacher have access to the reference solution; the student does not.

Your task is to read the student reasoning transcript and decide whether to intervene, and if so, when and what you would say. You may only intervene once during the student's reasoning process and before the student submits their final answer. If you choose to intervene, your intervention message will be sent to the student. You may also choose not to intervene.

Your response must follow this format. Do not include any explanations or additional text.

Intervene: [Yes/No]

Time: <If 'Yes', provide the 5 words in the reasoning transcript immediately preceding the point where you would intervene in the reasoning transcript (not in the student's final answer). Format as a single string: "word1 word2 word3 word4 word5"
If 'No', leave this blank.>

Intervention: <If 'Yes', write your intervention message to the student at this moment. If 'No', leave this blank.>

(*@\textbf{[User]}@*)
Title: {{ question_title }}
Problem: {{ question_content }}

Student reasoning: {{ full_transcript }}

Student final answer: {{ student_final_answer }}

Reference solution: {{ reference_content }}

Evaluation: {{ evaluation }}
\end{lstlisting}
\end{tcolorbox}

\clearpage
\subsection{Post-Intervention Student Reasoning}
If a teacher decides to intervene, the student updates their reasoning. We present the prompt for continuing the reasoning process below, followed by the prompt for the \textit{Stop-and-Answer} ablation.

\begin{tcolorbox}[compactbox, title=Post-Intervention Student Reasoning Prompt, width=\textwidth]
\begin{lstlisting}[breaklines=true, basicstyle=\footnotesize\ttfamily, escapeinside={(*@}{@*)}]
(*@\textbf{[System]}@*)

[Student baseline system prompt]

During your reasoning process, a teacher may intervene and provide updates in the format: <update>...</update>. Please incorporate the teacher's update into your reasoning process.

(*@\textbf{[User 1]}@*)
Title: {{ question_title }}
Problem: {{ question_content }}

(*@\textbf{[Assistant]}@*)
Reasoning: {{ reasoning_snippet }}

(*@\textbf{[User 2]}@*)
<update>{{ teacher_intervention }}</update>

Please continue your reasoning from where you left off, incorporating the teacher's feedback, and provide your final answer.
\end{lstlisting}
\end{tcolorbox}

In the \textit{Stop-and-Answer} condition, the second user prompt is instead replaced with: ``Please incorporate the teacher's feedback and provide your final answer immediately using the format specified in the system prompt.''

\subsection{Judge}
\label{sec:judge-prompt}

\begin{tcolorbox}[compactbox, title=Judge Prompt, width=\textwidth]
\begin{lstlisting}[breaklines=true, basicstyle=\footnotesize\ttfamily, escapeinside={(*@}{@*)}]
(*@\textbf{[System]}@*)
You are an expert evaluator. You will be given a brain teaser and its solution. Your task is to determine whether the candidate solution is correct.

Your response must follow this format. Do not include any additional text.

Verdict: Correct/Incorrect

Explanation: <One sentence explaining why correct or incorrect.>


(*@\textbf{[User]}@*)
Title: {{ question_title }}
Problem: {{ question_content }}

Candidate solution: {{ student_final_answer }}

Reference solution: {{ reference_content }}
\end{lstlisting}
\end{tcolorbox}

\clearpage
\subsection{Problem Generation}
\label{sec:prompts-generation}

\begin{tcolorbox}[compactbox, title=Skill Extractor $\mathcal{E}$ Prompt, width=\textwidth]
\begin{lstlisting}[breaklines=true, basicstyle=\footnotesize\ttfamily, escapeinside={(*@}{@*)}]
Your task is to label the following [DOMAIN] problem with a [SKILL_TYPE] skill that a student would need to correctly [TASK_DESCRIPTION].

# Rules
- The skill name should be usable as a dictionary key in Python.
- The skill name should use lowercase letters only.
- The skill name should be very descriptive and may use multiple words to describe the [SKILL_TYPE] skills required.
- If you use multiple words, join them with underscores.

# Problem
[PROBLEM_FIELDS]

# Output format
Your response must follow this format:
<name_of_the_skill>, reason: <reason_for_the_skill>
\end{lstlisting}
\end{tcolorbox}

\begin{tcolorbox}[compactbox, title=Skill Clusterer $\mathcal{C}$ Prompt, width=\textwidth]
\begin{lstlisting}[breaklines=true, basicstyle=\footnotesize\ttfamily, escapeinside={(*@}{@*)}]
Here is a list of skills required to solve a [DOMAIN] problem:
{skills_list}

Reduce the number of unique skills by grouping similar skills into categories and give a descriptive name to each category.
\end{lstlisting}
\end{tcolorbox}

\begin{tcolorbox}[compactbox, title=Problem Generator $\mathcal{G}$ Prompt, width=\textwidth]
\begin{lstlisting}[breaklines=true, basicstyle=\footnotesize\ttfamily, escapeinside={(*@}{@*)}]
You are given a reference [DOMAIN] problem. Your task is to generate a new [DOMAIN] problem that tests the same underlying skill, while being meaningfully different in surface form.

# Reference Problem

[REFERENCE_PROBLEM_FIELDS]

Target Skill (do not change):
{skill_name}
{reason}

# Instructions
Generate a new [DOMAIN] problem that satisfies all of the following:
[DOMAIN_SPECIFIC_REQUIREMENTS]

# Output format
Your response must follow this format:
[OUTPUT_FORMAT_FIELDS]
\end{lstlisting}
\end{tcolorbox}

\begin{tcolorbox}[compactbox, title=Validator $\mathcal{V}$ Prompt, width=\textwidth]
\begin{lstlisting}[breaklines=true, basicstyle=\footnotesize\ttfamily, escapeinside={(*@}{@*)}]
You are an expert [DOMAIN] evaluator. Your task is to verify if the given [SOLUTION_TYPE] correctly solves the problem.

[PROBLEM_FIELDS]

[SOLUTION_FIELDS]

Does this [SOLUTION_TYPE] correctly solve the problem described above?

# Output format
[OUTPUT_FORMAT]
\end{lstlisting}
\end{tcolorbox}

\subsection{Generalization Evaluation}
We evaluate the student's ability to generalize to new problems under two different context conditions.
Both conditions use the same system prompt as \secref{sec:student-prompts}; we show only the user message turns below.

\begin{tcolorbox}[compactbox, title=Intervention-Context Condition Student Prompt, width=\textwidth]
% (*@\textbf{[User]}@*)
\begin{lstlisting}[breaklines=true, basicstyle=\footnotesize\ttfamily, escapeinside={(*@}{@*)}]
You previously worked on the following problem:

Title: {{ prev_question_title }}
Problem: {{ prev_question_content }}

Your initial reasoning process:
{{ prev_reasoning_snippet }}

Your teacher intervened at this point with the following feedback:
{{ prev_intervention }}

Your revised reasoning based on the teacher's feedback:
{{ prev_counterfactual_reasoning }}

Your final answer:
{{ prev_counterfactual_answer }}

Evaluation: {{ prev_judge_verdict }}
Explanation: {{ prev_judge_explanation }}


Now, solve the following problem:

Title: {{ current_question_title }}
Problem:
{{ current_question_content }}

Use the lessons learned from the previous problem and your teacher's feedback to help you solve this problem.
\end{lstlisting}
\end{tcolorbox}

\begin{tcolorbox}[compactbox, title=Problem-Context Condition Student Prompt, width=\textwidth]
% (*@\textbf{[User]}@*)
\begin{lstlisting}[breaklines=true, basicstyle=\footnotesize\ttfamily, escapeinside={(*@}{@*)}]
You previously worked on the following problem:

Title: {{ prev_question_title }}
Problem: {{ prev_question_content }}

Your reasoning process:
{{ prev_reasoning_trace }}

Your final answer:
{{ prev_final_answer }}

Evaluation: {{ prev_judge_verdict }}
Explanation: {{ prev_judge_justification }}

Now, your task is to solve the following problem:

Title: {{ current_question_title }}
Problem:
{{ current_question_content }}
\end{lstlisting}
\end{tcolorbox}

\clearpage
\subsection{Intervention Categorization}
\label{sec:prompts-categorization}

The functional role and solution leakage labeling prompts share the same context block; only the task-specific section (\texttt{\# Your Task}) differs. 
The shared context is:

\begin{tcolorbox}[compactbox, title=Shared Context, width=\textwidth]
\begin{lstlisting}[breaklines=true, basicstyle=\footnotesize\ttfamily]
# Problem Context
Title: {title}
Reference solution: {reference_solution}

# Student Reasoning Process
The student was working through the problem step by step. The reasoning is shown in chunks of 50 characters each. Here is their reasoning up to the point where the LLM intervened:

Intervention occurred at chunk: {intervention_at_chunk}
Chunks shown before intervention: {chunks_shown}
Total chunks in reasoning: {total_chunks}

Student reasoning up to intervention point:
{reasoning_up_to_intervention}

# Intervention Details
LLM intervention message:
{intervention_message}
\end{lstlisting}
\end{tcolorbox}

\begin{tcolorbox}[compactbox, title=Functional Role Task Prompt, width=\textwidth]
\begin{lstlisting}[breaklines=true, basicstyle=\footnotesize\ttfamily, escapeinside={(*@}{@*)}]
You are analyzing an LLM teaching intervention scenario. An LLM teacher intervened at a specific point during a student's problem-solving process. Your task is to understand how this intervention is helping the student.

(*@\textbf{[Shared Context]}@*)

# Your Task
Based on the student's reasoning up to the intervention point and the LLM's intervention message, provide a concise one-line explanation for how this intervention is helping the student at this specific point in their reasoning process.

Your response must follow this format:
Reason: <your one-line explanation of how the intervention helps the student>
\end{lstlisting}
\end{tcolorbox}

\begin{tcolorbox}[compactbox, title=Solution Leakage Task Prompt, width=\textwidth]
\begin{lstlisting}[breaklines=true, basicstyle=\footnotesize\ttfamily, escapeinside={(*@}{@*)}]
You are analyzing an LLM teaching intervention to determine how much of the solution it reveals to the student.

(*@\textbf{[Shared Context]}@*)

# Your Task
Compare the intervention message to the reference solution above and analyze how much of the solution is revealed in the intervention.

Your response must follow this format:
Solution Revealed: <your one-line description of how much solution is revealed>
\end{lstlisting}
\end{tcolorbox}

\clearpage
\begin{tcolorbox}[compactbox, title=Functional Role Clustering Prompt, width=\textwidth]
\begin{lstlisting}[breaklines=true, basicstyle=\footnotesize\ttfamily]
Here is a list of reasons explaining how interventions (from both human and LLM teachers) are helping students during their problem-solving processes:
{reasons_str}

These reasons explain how interventions help students at specific points in their reasoning. Reduce the number of unique reasons by grouping similar reasons into high-level categories and give a descriptive name to each category.

Focus on the underlying type of help or learning support provided rather than surface-level differences in wording.

IMPORTANT: Ensure the categories are NOT domain-specific. Do not mention specific subjects like math, coding, or brain teasers. The categories should apply generally to teaching interventions across any domain.
\end{lstlisting}
\end{tcolorbox}

\begin{tcolorbox}[compactbox, title=Solution Leakage Clustering Prompt, width=\textwidth]
\begin{lstlisting}[breaklines=true, basicstyle=\footnotesize\ttfamily]
Here is a list of descriptions explaining how much of the solution interventions (from both human and LLM teachers) reveal to students during their problem-solving processes:
{reasons_str}

These descriptions explain how much of the solution is revealed in each intervention, focusing on which solution steps or key insights are disclosed. Reduce the number of unique descriptions by grouping similar descriptions into high-level categories and give a descriptive name to each category.

Focus on the underlying level or type of solution revelation rather than surface-level differences in wording.

IMPORTANT: Ensure the categories are NOT domain-specific. Do not mention specific subjects like math, coding, or brain teasers. The categories should apply generally to teaching interventions across any domain.
\end{lstlisting}
\end{tcolorbox}

\begin{tcolorbox}[compactbox, title=Relabeling Prompt, width=\textwidth]
\begin{lstlisting}[breaklines=true, basicstyle=\footnotesize\ttfamily]
You are categorizing an intervention reason into one of the existing high-level categories.

**Intervention reason to categorize:**
{reason}

**Available categories:**
{categories_text}

Choose the ONE category from the list above that best matches this intervention reason.

Your response must follow this format:
Category: <exact category name from the list above>
\end{lstlisting}
\end{tcolorbox}

\clearpage
\subsection{Prompt Variations}
Both variations share the \textit{Standard} teacher's increment mechanics, output format, and user message (\secref{sec:teacher-prompts}):

\begin{tcolorbox}[compactbox, title=Prompt Variation Teacher Prompt, width=\textwidth]
\begin{lstlisting}[breaklines=true, basicstyle=\footnotesize\ttfamily, escapeinside={(*@}{@*)}]
(*@\textbf{[System]}@*)
You are watching a user attempt a code debugging problem.

(*@\textbf{[Variation 2]}@*)

You will see the reasoning in increments of {{ chunk_size }} characters at a time. After each increment, you will be asked whether you would like to intervene. If you choose not to intervene, you will then receive the next {{ chunk_size }} characters, along with all previously shown text. You may only intervene once. If you choose to intervene, the task ends immediately and your intervention message will be sent to the student. (*@\textbf{[Variation~1]}@*) You may also choose not to intervene.

Your response must follow this format. Do not include any explanations or additional text.

Intervene: [Yes/No]

Intervention: <If 'Yes', write your intervention message to the student at this moment. If 'No', leave this blank.>


(*@\textbf{[User]}@*)
Title: {{ question_title }}

Description: {{ question_content }}

Buggy code: {{ source_text }}

Student reasoning: {{ transcript_portion }}

Reference solution: {{ reference_content }}

Do you want to intervene?
\end{lstlisting}
\end{tcolorbox}

For Variation 1, the \texttt{[Variation 1]} section is: ``Only intervene if truly necessary.''

The \texttt{[Variation 2]} section includes the following text: ``Your goal is to enable the student to reason independently and learn, not simply help the student get the correct answer quickly. You should only intervene when absolutely necessary, and your message should scaffold and guide the student.''

\end{document}